\documentclass{article}

     \usepackage[preprint]{NeurIPS_template/neurips_2025}

\usepackage[utf8]{inputenc} 
\usepackage[T1]{fontenc}    
\usepackage{hyperref}       
\usepackage{url}            
\usepackage{booktabs}       
\usepackage{amsfonts}       
\usepackage{nicefrac}       
\usepackage{microtype}      
\usepackage{xcolor}         

\usepackage{mathbbol}
\usepackage{thmtools,thm-restate}

\usepackage{tikz} 
\usepackage{amsmath,amsfonts,amssymb,dsfont,bm,amsthm}
\usepackage{csquotes}
\usepackage{mathbbol}
\usepackage{subcaption}
\usepackage{thmtools,thm-restate}
\usepackage{MnSymbol}
\usepackage{bm}
\usepackage{accents}

\usepackage{ciphod}

 \newtheorem{definition}{Definition}
 \newtheorem{theorem}{Theorem}
 \newtheorem{example}{Example}

\newcommand{\Myack}[1]{\begin{ack}#1\end{ack}}

\title{Identifying Macro Causal Effects in C-DMGs over DMGs}

\author{%
  Simon Ferreira \\
Sorbonne Université, INSERM, \\Institut Pierre Louis d’Epidémiologie\\ et de Santé Publique,\\
F75012, Paris, France\\
    \texttt{simon.ferreira@sorbonne-universite.fr} \\
  \And
  Charles K. Assaad \\
Sorbonne Université, INSERM, \\Institut Pierre Louis d’Epidémiologie\\ et de Santé Publique,\\
F75012, Paris, France\\
  \texttt{charles.assaad@inserm.fr} \\
}

\begin{document}

\maketitle

\begin{abstract}
The do-calculus is a sound and complete tool for identifying causal effects in acyclic directed mixed graphs (ADMGs) induced by structural causal models (SCMs).
However, in many real-world applications, especially in high-dimensional setting, constructing a fully specified ADMG is often infeasible.
This limitation has led to growing interest in partially specified causal representations, particularly through cluster-directed mixed graphs (C-DMGs), which group variables into clusters and offer a more abstract yet practical view of causal dependencies.
While these representations can include cycles, recent work has shown that the do-calculus remains sound and complete for identifying macro-level causal effects in C-DMGs over ADMGs under the assumption that all clusters size are greater than $1$.
Nevertheless, real-world systems often exhibit cyclic causal dynamics at the structural level.
To account for this, input-output structural causal models (ioSCMs) have been introduced as a generalization of SCMs that allow for cycles.
ioSCMs induce another type of graph structure known as a directed mixed graph (DMG).
Analogous to the ADMG setting, one can define C-DMGs over DMGs as high-level representations of causal relations among clusters of variables.
In this paper, we prove that, unlike in the ADMG setting, the do-calculus is unconditionally sound and complete for identifying macro causal effects in C-DMGs over DMGs.
Furthermore, we show that the graphical criteria for non-identifiability of macro causal effects previously established C-DMGs over ADMGs  naturally extends to a subset of C-DMGs over DMGs.
\end{abstract}

\section{Introduction}
Understanding and identifying causal effects is a central goal in many scientific disciplines.
In recent years, structural causal models (SCMs) have emerged as a foundational framework for reasoning about causality.
These models encode causal assumptions through structural equations and are typically represented by acyclic directed mixed graphs (ADMGs), which capture both causal and confounding relationships.
Within this framework, the do-calculus~\citep{Pearl_1995}\textemdash based on the notion of d-separation~\citep{Pearl_1988}\textemdash provides a complete and sound set of inference rules for identifying causal effects from observational data, assuming the causal structure is fully specified.
However, SCMs do not fully capture systems with cyclic causal dependencies at the structural level, which are common in public health, biology, economics, and engineering systems.
For example, there can be  a cyclic relation between
poor mental health (\eg, depression or anxiety) and substance use (\eg, alcohol, drugs).
The worsening of mental health and increase in substance use can occur in tight time-frames (daily or even hourly), especially in high-risk populations. Over time, they may reach a cyclic equilibrium where both reinforce each other without a clear causal ordering.
To address this, the notion of input-output structural causal models (ioSCMs) has been proposed~\citep{Forre_2020}.
These models generalize SCMs by allowing for cycles and induce a new class of graphs known as directed mixed graphs (DMGs)~\citep{Richardson_1997, Forre_2017, Forre_2018, Forre_2020,Boeken_2024}, which provide a richer representation of causal structures.
Furthermore, \cite{Forre_2020} introduced an extension of d-separation to DMGs, called $\sigma$-separation, and showed that the do-calculus, when replacing d-separation by $\sigma$-separation  becomes sound for identifying causal effects in DMGs~\citep{Forre_2020}.

However, in many real-world applications\textemdash particularly those involving high-dimensional data or limited domain knowledge\textemdash it is often unrealistic to assume a complete specification of the underlying causal graph.
This has motivated the development of partially specified graphical models~\citep{Maathuis_2013, Perkovic_2016,Perkovic_2020,Jaber_2022, Wang_2023,Assaad_2023,Anand_2023,Wahl_2024,Reiter_2024,Boeken_2024,Ferreira_2024,Ferreira_2025}, and in particular cluster graphs.
Cluster graphs abstract away some of the fine-grained details by grouping variables into clusters, thus offering a more flexible and scalable representation of complex systems. 
Importantly, cluster graphs allow for cycles, which can arise naturally in feedback systems or time-dependent processes, complicating the analysis compared to traditional ADMGs.
In these graphs, causal effects can be separated into two types: a micro causal effect where the interest is the effect of variable within a cluster on another variable in another cluster; and the macro causal effect where the interest in the effect of a set of an entire cluster on another entire cluster.
In this work, we focus on the latter.
\cite{Anand_2023,Tikka_2023} have shown that do-calculus (the version using d-separation) remains both sound and complete for identifying macro causal effects when the cluster graph representing ADMGs is acyclic.
\cite{Ferreira_2025,Ferreira_2025b} showed that the do-calculus (the version using d-separation) is also sound and complete for identifying macro-level causal effects when the cluster graph representing an ADMG is cyclic, denoted here as C-DMG over ADMGs, assuming either that the size of the clusters is unknown, or that each cluster contains more than one variable~\citep{Ferreira_2025b}.


Motivated by these developments, we consider the problem of identifying macro-level causal effects in cluster graphs representing DMGs, denoted as C-DMGs over DMGs, a natural generalization of previous work. Our contributions are threefold:

\begin{itemize}
\item We prove that  $\sigma$-separation~\citep{Forre_2018}\textemdash a fundamental tool in causal reasoning in DMGs\textemdash is sound and complete in C-DMGs over DMGs.
\item We prove that do-calculus (the version using $\sigma$-separation)~\citep{Forre_2020} is sound and complete for identifying macro-level causal effects in C-DMGs over DMGs\textemdash unconditionally, and without the constraints needed in the case of C-DMGs over ADMGs~\citep{Ferreira_2025b,Yvernes_2025}.
\item We show that the graphical characterization of non-identifiability previously developed for C-DMGs over ADMGs~\citep{Ferreira_2025,Ferreira_2025b} also applies for C-DMGs over DMGs under an additional assumption.
\end{itemize}

The remainder of the paper is organized as follows: In Section~\ref{sec:notations}, we formally presents C-DMGs over DMGs.
In Section~\ref{sec:macro}, we show that $\sigma$-separation and the do-calculus is sound and complete for macro causal effects in C-DMGs over DMGs and present a graphical characterization for the non-identifiability of these effects.
Finally in Section~\ref{sec:conclusion}, we conclude the paper while showing its limitations.
All proofs are deferred to the appendix.

\section{Preliminaries}
\label{sec:notations}

To streamline the presentation and avoid repetitive explanations, we will adopt the unified notation $\mathcal{G}^* = (\mathbb{V}^*, \mathbb{E}^*)$ to refer to any type of graph. This notation allows us to generalize results and discussions without redundancy across different graph types.
In the remainder, for every vertex $V^* \in \mathbb{V}^*$ in a graph $\mathcal{G}^*=(\mathbb{V}^*,\mathbb{E}^*)$, 
we will refer to its parents in graph by, $\parents{V^*}{\mathcal{G}^*}$, its ancestors by $\ancestors{V^*}{\mathcal{G}^*}$, and its descendants by $\descendants{V^*}{\mathcal{G}^*}$. We consider that a vertex counts as its own descendant and as its own ancestor. 
In addition, the strongly connected component of a vertex is defined as $\scc{V}{\mathcal{G}^*} = \ancestors{V}{\mathcal{G}^*} \cap \descendants{V}{\mathcal{G}^*}$.

In this section, we present the essential definitions and notations that will be used throughout the paper, ensuring clarity and consistency in the exposition of our results. In this work, we assume causal relations are modeled using an input/output structural causal model (ioSCM)~\citep{Forre_2020}\textemdash which extends classical structural causal models (SCMs)~\citep{Pearl_2000} by allowing for the presence of cycles.
Unlike classical SCMs, ioSCMs allow structural equations to mutually depend on each other.
For example, in the cyclic system:
\begin{equation*}
    X:=f_X(Y,L_X)   \qquad ; \qquad   Y:=f_Y(X,L_Y)   \qquad ; \qquad   (X,Y):=f_{(X,Y)}(L_X,L_Y),
\end{equation*}
$X$ functionally depends on $Y$, and $Y$ functionally depends on $X$, forming  cycle.
When cycles exist, instead of computing variables in a top-down order as in SCMs, ioSCMs rely on fixed-point solutions. That is, a joint assignment to the variables that simultaneously satisfies all equations. This is analogous to finding an equilibrium in dynamic systems.
Firstly, let us properly define the notion of loops as it will be useful to guarantee the compatibility of the causal mechanisms in  ioSCMs.
\begin{definition}[Loops]
    In a directed graph  $\mathcal{G}^*=(\mathbb{V}^*, \mathbb{E}^*)$, a loop is a set of vertices $\mathbb{S}\subseteq\mathbb{V}^*$ such that there exists a directed path between every pair of distinct vertices in the subgraph induced by $\mathbb{S}$ \ie, $\forall U \neq V \in \mathbb{S}, V\in\descendants{U}{ \inducedsthg{\mathcal{G}^*}{\mathbb{S}}}$.

    Note that every singleton $\mathbb{S}\in\{\{V\}\mid V\in\mathbb{V}^*\}$ and every strongly connected components $\mathbb{S}\in\{\scc{V}{\mathcal{G}^*}\mid V\in\mathbb{V}^*\}$ are loops.
    We write the set of all loops of the graph $\mathcal{G}^*$ as $\loops{\mathcal{G}^*}$. We call cycles the loops that are not singletons.
\end{definition}

Next, we recall the definition of ioSCMs from~\cite{Forre_2020} with the omission of the domains of the variables.
\begin{definition}[input/output Structural Causal Model (ioSCM)]
    An input/output structural causal model is a tuple $\mathcal{M}=(\latentset, \observedset, \intervenedset, \mathcal{G}^+, \mathbb{F},  \proba{\latentval})$,
	where
    \begin{itemize}
        \item $\latentset$ is a set of latent/exogenous variables, which cannot be observed but affect the rest of the model.
        \item $\observedset$ is a set of observed/endogenous variables, which are observed and every $V \in \observedset$ is functionally dependent on some subset of $\left(\latentset \cup \observedset \cup \intervenedset\right) \backslash \left\{V\right\}$.
        \item $\intervenedset$ is a set of input/intervention variables which are not functionally dependent of any other variable but rather are fixed to specific values.
        \item $\mathcal{G}^+=(\mathbb{V}^+,\mathbb{E}^+)$ is a graphical structure where:
        \begin{itemize}
            \item $\mathbb{V}^+ = \observedset \cup \latentset \cup \intervenedset$
            \item $\observedset = \children{\latentset\cup\intervenedset}{\mathcal{G}^+}$
            \item $\parents{\latentset\cup\intervenedset}{\mathcal{G}^+} = \emptyset$
        \end{itemize}
        \item $\mathbb{F}$ is a set of functions such that for all $\mathbb{S} \in \loops{\inducedsthg{\mathcal{G}^+}{\observedset}}$, $f^{\mathbb{S}}$ is a function taking as input the values of $\parents{\mathbb{S}}{\mathcal{G}^+}\backslash\mathbb{S}$ and outputting values for $\mathbb{S}$ and such that $\mathbb{F}$ satisfies the global compatibility condition:
        \begin{equation}
        \begin{gathered}
        \forall \mathbb{S}'\subsetneq\mathbb{S}\in\loops{\inducedsthg{\mathcal{G}^+}{\observedset}}, \forall \observedval\text{ values of }\parents{\mathbb{S}}{\mathcal{G}^+}\cup\mathbb{S},\\
        f^\mathbb{S}(\inducedsthg{\observedval}{{\parents{\mathbb{S}}{\mathcal{G}^+}\backslash\mathbb{S}}}) = \inducedsthg{\observedval}{{\mathbb{S}}}\implies f^{\mathbb{S}'}(\inducedsthg{\observedval}{{\parents{\mathbb{S}'}{\mathcal{G}^+}\backslash\mathbb{S}'}}) = \inducedsthg{\observedval}{{\mathbb{S}'}}
        \label{ass:compatibility_of_causal_mechanisms}
        \end{gathered}
        \end{equation}
        \item 	$\proba{\latentval}$ is a joint probability distribution over $\latentset$.
    \end{itemize}
\end{definition}

An ioSCM induces a directed graph, where every variable in $\mathbb{V}^+$ corresponds to a vertex in the graph. In this directed graph, a directed edge $\rightarrow$ is drawn from one variable to another if the former serves as an input to the function that determines the latter. For simplicity, instead of working directly with these directed graphs, we consider an alternative representation known as an  directed mixed graph (DMG). In a DMG, only the observed and intervened variables (\ie, $\observedset\cup\intervenedset$) correspond to vertices, while hidden variables in $\mathbb{L}$ that share common outputs are represented by bidirected edges $\longdashleftrightarrow$ between the corresponding observed variables, thereby implicitly accounting for the hidden confounding. 
Formally, DMGs are defined as follows:


\begin{definition}[Directed mixed graph (DMG)]
    \label{def:ADMG}
    Consider an ioSCM $\mathcal{M}$. The directed mixed graph $\mathcal{G}=(\mathbb{V},\mathbb{E}=\mathbb{E}_{\rightarrow} \cup \mathbb{E}_{\longdashleftrightarrow})$ induced by $\mathcal{M}$ is the DMG where:
    \begin{itemize}
        \item the vertices $\mathbb{V}=\observedset\cup\intervenedset$ are the endogenous variables and the intervention variables of the ioSCM; and
        \item the directed edges in $\mathcal{G}$ are $\mathbb{E}_\rightarrow=\inducedsthg{\mathbb{E}^+}{\observedset\cup\intervenedset}$; and
        \item the bidirected edges in $\mathcal{G}$ are $\mathbb{E}_{\longdashleftrightarrow} = \left\{X \longdashleftrightarrow Y\mid X,Y\in \mathbb{V}, \exists L\in\latentset \st\right.$ $\left. L\rightarrow X,L\rightarrow Y \in \mathbb{E}^+\right\}$.
    \end{itemize} 
\end{definition}

However, in many fields, constructing, analyzing, and validating a DMG remains a significant challenge for researchers due to the inherent difficulty in accurately determining causal relationships among individual variables. This complexity primarily stems from the uncertainty surrounding causal relations, making it challenging to specify the precise structure of the graph.
Nevertheless, researchers can often provide a partially specified version of the DMG, which offers a more practical and compact representation of the underlying causal structure. These simplified representations, which we call Cluster-Directed Mixed Graphs over DMGs (C-DMGs over DMGs), group several variables into clusters, allowing for the representation of causal relationships at a higher level of abstraction while retaining essential structural properties of the system. In a C-DMG over DMGs, directed edges between clusters represent causal influences at the higher level, while bidirected edges capture hidden confounding effects that exist between clusters.
Formally, C-DMGs over DMGs are defined as follows:

\begin{definition}[Cluster directed mixed graph over DMGs (C-DMG over DMGs)]
    \label{def:C-DMG}
    Let $\mathcal{G}=(\mathbb{V},\mathbb{E})$ be a DMG induced from an ioSCM $\mathcal{M}$. A C-DMG over DMGs is a graph $\mathcal{G}^\mathbb{c}=(\mathbb{C},\mathbb{E}^\mathbb{c})$ where: 
    \begin{itemize}
        \item $\mathbb{C}=\{{C}_1,\cdots,{C}_k\}$ is a partition (\ie, disjoints sets of micro-variables) of $\mathbb{V}$; and
        \item  $\forall {C}_i,{C}_j \in \mathbb{C}$ the edge ${C}_i\rightarrow {C}_j$ (resp. ${C}_i\longdashleftrightarrow {C}_j$) is in $\mathbb{E}^\mathbb{c}$ if and only if there exists $V_i\in {C}_i$ and $V_j \in {C}_j$ such that $V_i \rightarrow V_j$ (resp. $V_i \longdashleftrightarrow V_j$) is in $\mathbb{E}$.
    \end{itemize} 
\end{definition}

\begin{figure*}[t!]
	\centering
	\begin{subfigure}{.33\textwidth}
		\centering
				\begin{tikzpicture}[{black, circle, draw, inner sep=0}]
			\tikzset{nodes={draw,rounded corners},minimum height=0.7cm,minimum width=0.6cm, font=\scriptsize}
			\tikzset{latent/.append style={white, fill=black}}
			
			\node  (W1) at (-0.5,-2) {$W_1$};
			\node (W2) at (0.5,-2) {$W_2$};
			\node  (W3) at (-1,-3) {$W_3$};
			\node  (W4) at (1,-3) {$W_4$};
			\node[fill=red!30] (X1) at (-1,0) {$X_1$};
			\node[fill=red!30] (X2) at (-1.5,-1) {$X_2$};
			\node[fill=red!30]  (X3) at (-0.5,-1) {$X_3$};
			\node[fill=blue!30] (Y1) at (1,0) {$Y_1$};
			\node[fill=blue!30] (Y2) at (1.5,-1) {$Y_2$};

                \draw[->,>=latex] (X1) to (X2);
                \draw[->,>=latex] (X1) to (X3);

                \draw[->,>=latex] (W3) to (W2);
                \draw[->,>=latex] (W4) to (W1);

			\begin{scope}[transform canvas={yshift=-.15em}]
				\draw [->,>=latex] (X2) -- (X3);
			\end{scope}
			\begin{scope}[transform canvas={yshift=.15em}]
				\draw [<-,>=latex] (X2) -- (X3);
			\end{scope}		
            
			\begin{scope}[transform canvas={xshift=-.15em}]
				\draw [->,>=latex] (X2) -- (W3);
			\end{scope}
			\begin{scope}[transform canvas={xshift=.15em}]
				\draw [<-,>=latex] (X2) -- (W3);
			\end{scope}			
                \draw[->,>=latex] (Y2) to (W4);
                \draw[->,>=latex] (W2) to (Y1);

    	\draw[<->,>=latex, dashed] (W3) to [out=-80,in=-100, looseness=1] (W4);
            \draw[<->,>=latex, dashed] (X1) to [out=80,in=100, looseness=1] (Y1);

		\end{tikzpicture}		
        \caption{DMG1}
		\label{fig:DMGs_CDMG:DMG1}
	\end{subfigure}
	\hfill 
	\begin{subfigure}{.33\textwidth}
		\centering
		\begin{tikzpicture}[{black, circle, draw, inner sep=0}]
			\tikzset{nodes={draw,rounded corners},minimum height=0.7cm,minimum width=0.6cm, font=\scriptsize}
			\tikzset{latent/.append style={white, fill=black}}
			
			\node  (W1) at (-0.5,-2) {$W_1$};
			\node (W2) at (0.5,-2) {$W_2$};
			\node  (W3) at (-1,-3) {$W_3$};
			\node  (W4) at (1,-3) {$W_4$};
			\node[fill=red!30] (X1) at (-1,0) {$X_1$};
			\node[fill=red!30] (X2) at (-1.5,-1) {$X_2$};
			\node[fill=red!30]  (X3) at (-0.5,-1) {$X_3$};
			\node[fill=blue!30] (Y1) at (1,0) {$Y_1$};
			\node[fill=blue!30] (Y2) at (1.5,-1) {$Y_2$};

                \draw[->,>=latex] (X1) to (X2);
                \draw[->,>=latex] (X3) to (X1);
                \draw[->,>=latex] (X2) to (X3);

                \draw[->,>=latex] (W1) to (W3);
                \draw[->,>=latex] (W2) to (W4);

                \draw[->,>=latex] (X2) to (W1);
                \draw[->,>=latex] (W1) to (X3);
                \draw[->,>=latex] (Y2) to (W2);
                \draw[->,>=latex] (W2) to (Y1);

    	\draw[<->,>=latex, dashed] (W1) to [out=-80,in=-100, looseness=1] (W2);
            \draw[<->,>=latex, dashed] (X1) to [out=80,in=100, looseness=1] (Y1);

		\end{tikzpicture}		
        \caption{\centering DMG 2.}
		\label{fig:DMGs_CDMG:DMG2}
	\end{subfigure}
    \hfill 
    	\begin{subfigure}{.23\textwidth}
    		\begin{tikzpicture}[{black, circle, draw, inner sep=0}]
			\tikzset{nodes={draw,rounded corners},minimum height=0.6cm,minimum width=0.6cm}	
			\tikzset{latent/.append style={white, fill=black}}
			
			\node[fill=red!30] (X) at (0,0) {$C_\mathbb{X}$} ;
			\node[fill=blue!30] (Y) at (2.4,0) {$C_\mathbb{Y}$};
			\node (Z) at (1.2,-1) {$C_\mathbb{W}$};
			
			\begin{scope}[transform canvas={yshift=-.25em}]
				\draw [->,>=latex] (X) -- (Z);
			\end{scope}
			\begin{scope}[transform canvas={yshift=.25em}]
				\draw [<-,>=latex] (X) -- (Z);
			\end{scope}			
			
			\begin{scope}[transform canvas={yshift=-.25em}]
				\draw [->,>=latex] (Z) -- (Y);
			\end{scope}
			\begin{scope}[transform canvas={yshift=.25em}]
				\draw [<-,>=latex] (Z) -- (Y);
			\end{scope}			
	
			\draw[<->,>=latex, dashed] (X) to [out=90,in=90, looseness=1] (Y);
			
			\draw[->,>=latex] (X) to [out=165,in=120, looseness=2] (X);
			\draw[->,>=latex] (Z) to [out=205,in=250, looseness=2] (Z);
			
			\draw[<->,>=latex, dashed] (Z) to [out=-25,in=-70, looseness=2] (Z);			
		\end{tikzpicture}
		\caption{C-DMG.}
		\label{fig:DMGs_CDMG:CDMG}
	\end{subfigure}
	\caption{Two ADMGs and with their compatible C-DMG. Red vertices represent the exposures of interest in and blue vertices represent the outcome of interest.}
	\label{fig:DAG_ADMG_CDMG}
\end{figure*}
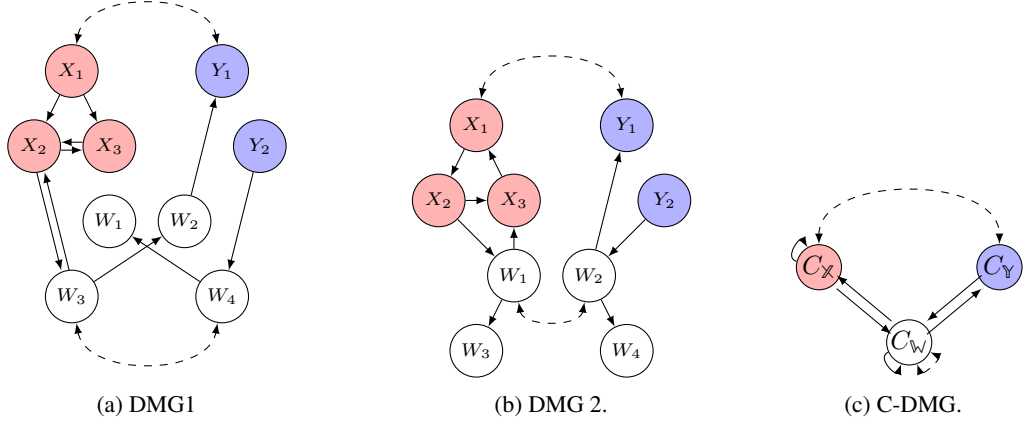

Figure~\ref{fig:DAG_ADMG_CDMG} presents a simple C-DMG over DMGs along with two of its compatible DMGs.
Cycles in a C-DMG over DMGs can arise for two distinct reasons. First, unlike in a C-DMG over ADMGs, a C-DMG over DMGs can contain a cycle if there is a genuine cyclic relationship in the underlying DMG between nodes belonging to different clusters.  For example, in Figure~\ref{fig:DMGs_CDMG:DMG1} the cycle between $X_2$ and $W_3$ in the DMG induces a cycle between clusters  $C_{\mathbb{X}}$ and $C_{\mathbb{W}}$ in the corresponding C-DMG over DMGs in Figure~\ref{fig:DMGs_CDMG:CDMG}.
Second, even in the absence of an actual cycle in the underlying DMG, cycles can appear in the C-DMG over DMGs due its partial specification. This is illustrated in Figure~\ref{fig:DMGs_CDMG:DMG2}, where the edges $X_2\rightarrow W_1$ and $W_1\rightarrow X_3$ together lead to  cycle at the between clusters $C_{\mathbb{X}}$ and $C_{\mathbb{W}}$ in the  C-DMG over DMGs in Figure~\ref{fig:DMGs_CDMG:CDMG}.
Lastly, cycles that are contained in a single cluster do not appear in the C-DMG over DMGs This is illustrated in Figure~\ref{fig:DMGs_CDMG:DMG1} with the cycle $X_2\rightleftarrows X_3$ that does not show in the C-DMG over DMGs in Figure~\ref{fig:DMGs_CDMG:CDMG}.

We distinguish between two types of causal effects in the context of C-DMGs, the macro causal effect~\citep{Anand_2023,Ferreira_2025,Ferreira_2025b} and the micro causal effect~\citep{Assaad_2024,Assaad_2025}. In this paper we focus on the former and we formally define it below:

\begin{definition}[Macro causal effect]
	\label{def:macro-total_effects}
    Consider a DMG $\mathcal{G}$ over variables $\mathbb{V}$ induced from an ioSCM and let $\mathcal{G}^\mathbb{c}=(\mathbb{C},\mathbb{E}^\mathbb{c})$ be a compatible C-DMG.
    A macro causal effect is a causal effect from a set of macro-variables $\mathbb{C}_\mathbb{X}$ on another set of macro-variables $\mathbb{C}_\mathbb{Y}$ where $\mathbb{C}_\mathbb{X}$ and $\mathbb{C}_\mathbb{Y}$ are disjoint subsets of $\mathbb{C}$.
    It is written $\probac{\mathbb{C}_\mathbb{Y} = \mathbb{c}_\mathbb{Y}}{\interv{\mathbb{C}_\mathbb{X} = \mathbb{c}_\mathbb{X}}}$, where the $\interv{\cdot}$ operator represents an external intervention.
\end{definition}

The identification problem in causal inference aims to establish whether a causal effect of a set of variables  on another set of variables can be expressed exclusively in terms of observed variables and standard probabilistic notions, such as conditional probabilities. 
Formally, the identification problem in the context of macro causal effects and C-DMGs over DMGs is defined as follows:
\begin{definition}[Identifiability in C-DMGs over DMGs]
\label{def:identification_problem}
Let $\mathbb{C}_\mathbb{X}$ and $\mathbb{C}_\mathbb{Y}$ be disjoint sets of vertices in a C-DMGs over DMGs $\mathcal{G}^{\mathbb{c}}$. The macro causal effect of $\mathbb{C}_\mathbb{X}$ on $\mathbb{C}_\mathbb{Y}$ is identifiable in
$\mathcal{G}^{\mathbb{c}}$ if $\probac{\mathbb{C}_\mathbb{Y} = \mathbb{c}_\mathbb{Y}}{\interv{\mathbb{C}_\mathbb{X} = \mathbb{c}_\mathbb{X}}}$ is uniquely computable from any observational positive distribution compatible with $\mathcal{G}^{\mathbb{c}}$.    
\end{definition}


In the following, we will abuse the notation by writing $\probac{\mathbb{c}_\mathbb{Y}}{\interv{\mathbb{c}_\mathbb{X}}}$ instead of $\probac{\mathbb{C}_\mathbb{Y} = \mathbb{c}_\mathbb{Y}}{\interv{\mathbb{C}_\mathbb{X} = \mathbb{c}_\mathbb{X}}}$ when the setting is clear. In addition, whenever the context is clear, we will refer to C-DMGs over DMGs simply as C-DMGs.

\section{Identification of Macro Causal Effects in C-DMGs over DMGs}
\label{sec:macro}

In this section, we aim to establish that the do-calculus is both sound and complete for identifying macro-level causal effects in a C-DMG over DMGs. We begin by showing, in the first subsection, that $\sigma$-separation—originally developed for DMGs as a tool for identifying conditional independencies—remains sound and complete when extended to C-DMGs over DMGs for detecting macro-level conditional independencies. In the second subsection, we present the core theoretical contribution of this section: the soundness and completeness of do-calculus for macro causal effect identification in this setting. Finally, we provide a graphical characterization of non-identifiability, shedding light on cases where causal effects cannot be inferred from observational data alone.

\subsection{The $\sigma$-separation in C-DMGs over DMGs}
The standard notion of d-separation~\citep{Pearl_1988} was originally introduced for acyclic directed mixed graphs (ADMGs). It was later shown to remain valid when extended to C-ADMGs over ADMGs~\citep{Anand_2023} and C-DMGs over ADMGs~\citep{Ferreira_2025,Ferreira_2025b}. However, d-separation does not apply to DMGs, which may contain true cyclic causal relations. To address this limitation, $\sigma$-separation was introduced as a generalization suitable for DMGs~\citep{Forre_2020}. In this subsection, we demonstrate that $\sigma$-separation can be naturally applied to C-DMGs over DMGs. We begin by formally defining  $\sigma$-blocked walks and the concept of $\sigma$-separation in this generalized setting.

\begin{definition}[$\sigma$-blocked walk \citep{Forre_2020}]
    \label{def:sigma_block}
    In a graph $\mathcal{G}^*=(\mathbb{V}^*,\mathbb{E}^*)$, a walk $\tilde{\pi}=\langle V^*_1,\cdots,V^*_n\rangle$ is said to be $\sigma$-blocked by a set of vertices $\mathbb{W}^*\subseteq\mathbb{V}^*$ if:
	\begin{enumerate}
        \item $V^*_1 \in \mathbb{W}^*$ or $V^*_n \in \mathbb{W}^*$, or
        
		\item $\exists 1 < i < n \st \langle V^*_{i-1}\stararrow V^*_{i}\arrowstar V^*_{i+1}\rangle \subseteq \tilde{\pi}$ and $V^*_{i}\notin\mathbb{W}^*$, or
        
		\item $\exists 1 < i < n \st \langle V^*_{i-1}\leftarrow V^*_{i}\arrowstar V^*_{i+1}\rangle \subseteq \tilde{\pi}$ and $V^*_{i}\in\mathbb{W}^*\backslash\scc{V^*_{i-1}}{\mathcal{G}^*}$, or
        
		\item $\exists 1 < i < n \st \langle V^*_{i-1}\stararrow V^*_{i}\rightarrow V^*_{i+1}\rangle \subseteq \tilde{\pi}$ and $V^*_{i}\in\mathbb{W}^*\backslash\scc{V^*_{i+1}}{\mathcal{G}^*}$, or

        \item $\exists 1 < i < n \st \langle V^*_{i-1}\leftarrow V^*_{i}\rightarrow V^*_{i+1}\rangle \subseteq \tilde{\pi}$ and \\ $V^*_{i}\in\mathbb{W}^*\backslash\left(\scc{V^*_{i-1}}{\mathcal{G}^*}\cap\scc{V^*_{i+1}}{\mathcal{G}^*}\right)$.

	\end{enumerate}
	where $\stararrow$ represents $\rightarrow$ or $\longdashleftrightarrow$, $\arrowstar$ represents $\leftarrow$ or $\longdashleftrightarrow$, and $\starbarstar$ represents any of the three arrow type $\rightarrow$, $\leftarrow$ or $\longdashleftrightarrow$.
	A walk which is not $\sigma$-blocked is said to be $\sigma$-active.
    
\end{definition}

\begin{definition}[$\sigma$-separation \citep{Forre_2020}]
	\label{def:s-separation}
    In a graph $\mathcal{G}^*=(\mathbb{V}^*,\mathbb{E}^*)$, let $\mathbb{X}^*,\mathbb{Y}^*, \mathbb{W}^*$ be distinct subsets of $\mathbb{V}^*$.
    $\mathbb{W}^*$ is said to $\sigma$-separate $\mathbb{X}^*$ and $\mathbb{Y}^*$ if and only if $\mathbb{W}^*$ $\sigma$-blocks every walk from a vertex in $\mathbb{X}^*$ to a vertex in $\mathbb{Y}^*$.
    It is written $\sigsepcgraph{\mathbb{X}^*}{\mathbb{Y}^*}{\mathbb{W}^*}{\mathcal{G}^*}$.
\end{definition}

The following theorem shows that $\sigma$-separation is applicable as is to C-DMGs over DMGs.

\begin{theorem}[Soundness of $\sigma$-separation in C-DMGs over DMGs]
	\label{theorem:soundness_s-sep}
	Let $\mathcal{G}^\mathbb{c}=(\mathbb{C},\mathbb{E}^\mathbb{c})$ be a C-DMG and $\mathbb{C}_\mathbb{X},\mathbb{C}_\mathbb{Y},\mathbb{C}_\mathbb{W}$ be disjoint subsets of $\mathbb{C}$.
    If $\mathbb{C}_\mathbb{X}$ and $\mathbb{C}_\mathbb{Y}$ are $\sigma$-separated by $\mathbb{C}_\mathbb{W}$ in $\mathcal{G}^\mathbb{c}$ then, in any compatible DMG $\mathcal{G}=(\mathbb{V},\mathbb{E})$, $\mathbb{X}=\bigcup_{C\in\mathbb{C}_\mathbb{X}}C$ and $\mathbb{Y}=\bigcup_{C\in\mathbb{C}_\mathbb{Y}}C$ are $\sigma$-separated by $\mathbb{W}=\bigcup_{C\in\mathbb{C}_\mathbb{W}}C$.
\end{theorem}

Theorem~\ref{theorem:soundness_s-sep} establishes that $\sigma$-separation in C-DMGs over DMGs ensures the existence of a corresponding macro-level $\sigma$-separation across all compatible DMGs. According to \cite[Theorem 5.2]{Forre_2020}, this implies that some conditional independencies in the underlying probability distribution can be inferred directly from the C-DMG. By extending the applicability of $\sigma$-separation to C-DMGs over DMGs, this result enables the identification of macro-level conditional independencies even when the underlying causal structure is only partially specified. To illustrate the practical value of this result, we now present two examples demonstrating the application of $\sigma$-separation in a C-DMG over DMGs.

\begin{example}
    Let $\mathcal{G}$ be the true \emph{unknown} DMG and consider  that its compatible C-DMG, denoted as $\mathcal{G}^\mathbb{c}$ is one given in in Figure~\ref{fig:identifiable_macro:a}. Using Definition~\ref{def:s-separation}, we can directly deduce $\sigsepcgraph{C_{\mathbb{W}}}{C_{\mathbb{Y}}}{C_{\mathbb{X}}}{\mathcal{G}^\mathbb{c}}$. Thus according to Theorem~\ref{theorem:soundness_s-sep} and \cite[Theorem 5.2]{Forre_2020}, $C_{\mathbb{W}}$ is conditionally independent of $C_{\mathbb{Y}}$ given $C_{\mathbb{X}}$ in every distribution compatible with the true ADMG.
\end{example}

\begin{example}
    Let $\mathcal{G}$ be the true \emph{unknown} DMG and consider  that its compatible C-DMG, denoted as $\mathcal{G}^\mathbb{c}$ is one given in in Figure~\ref{fig:identifiable_macro:c}. Using Definition~\ref{def:s-separation}, we can directly  deduce that $\sigsepcgraph{C_{\mathbb{Z}}}{C_{\mathbb{Y}}}{C_{\mathbb{X}},C_{\mathbb{W}}}{\mathcal{G}^\mathbb{c}}$. Thus according to Theorem~\ref{theorem:soundness_s-sep} and \cite[Theorem 5.2]{Forre_2020}, $C_{\mathbb{Z}}$ is conditionally independent of $C_{\mathbb{Y}}$ given $C_{\mathbb{X}}$ and $C_{\mathbb{W}}$ in every distribution compatible with the true ADMG.
\end{example}

The following theorem shows that $\sigma$-separation is also complete in C-DMGs over DMGs. 

\begin{theorem}[Completeness of $\sigma$-separation in C-DMGs]
	\label{theorem:completeness_s-sep}
	Let $\mathcal{G}^\mathbb{c}=(\mathbb{C},\mathbb{E}^\mathbb{c})$ be a C-DMG, $\mathbb{C}_\mathbb{X}, \mathbb{C}_\mathbb{Y}, \mathbb{C}_\mathbb{W}$ be disjoint subsets of $\mathbb{C}$, $\mathbb{X}=\bigcup_{C\in\mathbb{C}_\mathbb{X}}C$, $\mathbb{Y}=\bigcup_{C\in\mathbb{C}_\mathbb{Y}}C$ and $\mathbb{W}=\bigcup_{C\in\mathbb{C}_\mathbb{W}}C$.
    If $\mathbb{C}_\mathbb{X}$ and $\mathbb{C}_\mathbb{Y}$ are not $\sigma$-separated by $\mathbb{C}_\mathbb{W}$ in $\mathcal{G}^\mathbb{c}$, then there exists a compatible DMG $\mathcal{G}=(\mathbb{V},\mathbb{E})$ such that $\mathbb{X}$ and $\mathbb{Y}$ are not $\sigma$-separated by $\mathbb{W}$.
\end{theorem}

The findings of Theorems~\ref{theorem:soundness_s-sep} and~\ref{theorem:completeness_s-sep} establish that identifying a $\sigma$-separation in C-DMGs over DMGs ensures the recovery of all common macro-level $\sigma$-separations across all compatible DMGs. This result is particularly valuable for constraint-based causal discovery methods, especially when the goal is to infer the structure of a C-DMG without needing to fully specify an underlying DMG. Most importantly, these insights lay the theoretical groundwork for the results developed in the next subsection.

\subsection{The do-calculus in C-DMGs over DMGs}
\label{subsec:do-calc}
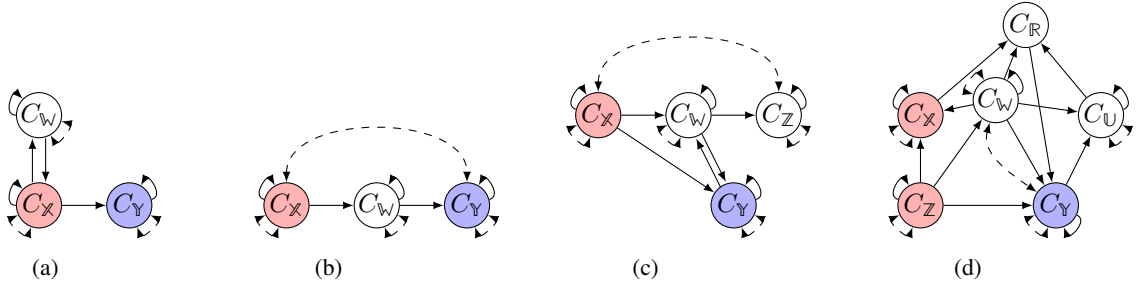
\begin{figure}[t!]
	\centering
	\begin{subfigure}{.09\textwidth}
		\centering
		\begin{tikzpicture}[{black, circle, draw, inner sep=0}]
			\tikzset{nodes={draw,rounded corners},minimum height=0.6cm,minimum width=0.6cm}	
			\tikzset{anomalous/.append style={fill=easyorange}}
			\tikzset{rc/.append style={fill=easyorange}}
			
			\node (W) at (0,1.2) {$C_{\mathbb{W}}$} ;
			\node[fill=red!30] (X) at (0,0) {$C_{\mathbb{X}}$} ;
			\node[fill=blue!30] (Y) at (1.2,0) {$C_{\mathbb{Y}}$};
			
			\draw [->,>=latex] (X) -- (Y);
			\begin{scope}[transform canvas={xshift=-.25em}]
				\draw [->,>=latex] (X) -- (W);
			\end{scope}
			\begin{scope}[transform canvas={xshift=.25em}]
				\draw [<-,>=latex] (X) -- (W);
			\end{scope}
			
			\draw[->,>=latex] (W) to [out=180,in=135, looseness=2] (W);
			\draw[->,>=latex] (X) to [out=180,in=135, looseness=2] (X);
			\draw[->,>=latex] (Y) to [out=15,in=60, looseness=2] (Y);
			
			\draw[<->,>=latex, dashed] (X) to [out=-155,in=-110, looseness=2] (X);
			\draw[<->,>=latex, dashed] (Y) to [out=-25,in=-70, looseness=2] (Y);
			\draw[<->,>=latex, dashed] (W) to [out=-15,in=-60, looseness=2] (W);	
		\end{tikzpicture}
		\caption{\centering}
		\label{fig:identifiable_macro:a}
	\end{subfigure}
	\hfill  
	\begin{subfigure}{.15\textwidth}
		\centering
		\begin{tikzpicture}[{black, circle, draw, inner sep=0}]
			\tikzset{nodes={draw,rounded corners},minimum height=0.6cm,minimum width=0.6cm}	
			\tikzset{latent/.append style={white, fill=black}}
			
			\node[fill=red!30] (X) at (0,0) {$C_{\mathbb{X}}$} ;
			\node[fill=blue!30] (Y) at (2.4,0) {$C_{\mathbb{Y}}$};
			\node (Z) at (1.2,0) {$C_{\mathbb{W}}$};

			\draw [->,>=latex,] (X) -- (Z);

			\draw[->,>=latex] (Z) -- (Y);

			\draw[<->,>=latex, dashed] (X) to [out=90,in=90, looseness=1] (Y);
			
			\draw[->,>=latex] (X) to [out=165,in=120, looseness=2] (X);
			\draw[->,>=latex] (Y) to [out=15,in=60, looseness=2] (Y);
			\draw[->,>=latex] (Z) to [out=15,in=60, looseness=2] (Z);

			\draw[<->,>=latex, dashed] (X) to [out=-155,in=-110, looseness=2] (X);
			\draw[<->,>=latex, dashed] (Y) to [out=-25,in=-70, looseness=2] (Y);
			\draw[<->,>=latex, dashed] (Z) to [out=-25,in=-70, looseness=2] (Z);			
			
		\end{tikzpicture}
		\caption{}
		\label{fig:identifiable_macro:b}
	\end{subfigure}
	\hfill 
	\begin{subfigure}{.16\textwidth}
		\centering
		\begin{tikzpicture}[{black, circle, draw, inner sep=0}]
			\tikzset{nodes={draw,rounded corners},minimum height=0.6cm,minimum width=0.6cm}	
			\tikzset{latent/.append style={white, fill=black}}
			
			\node[fill=red!30] (X) at (0,0) {$C_{\mathbb{X}}$} ;
			\node (Z) at (2.4,0) {$C_{\mathbb{Z}}$};
			\node (W) at (1.2,0) {$C_{\mathbb{W}}$};
			\node[fill=blue!30] (Y) at (1.8,-1.2) {$C_{\mathbb{Y}}$};

			\draw [->,>=latex,] (X) -- (Y);
			\draw [->,>=latex,] (X) -- (W);

			\draw[->,>=latex] (W) -- (Z);
               \begin{scope}[transform canvas={yshift=-.1em, xshift=-.1em}]
				\draw [->,>=latex] (Y) -- (W);
			\end{scope}
			\begin{scope}[transform canvas={yshift=.1em, xshift=.1em}]
				\draw [<-,>=latex] (Y) -- (W);
			\end{scope}

			\draw[<->,>=latex, dashed] (X) to [out=90,in=90, looseness=1] (Z);
			
			\draw[->,>=latex] (X) to [out=165,in=120, looseness=2] (X);
			\draw[->,>=latex] (Y) to [out=15,in=60, looseness=2] (Y);
			\draw[->,>=latex] (W) to [out=15,in=60, looseness=2] (W);
			\draw[->,>=latex] (Z) to [out=15,in=60, looseness=2] (Z);
			
			\draw[<->,>=latex, dashed] (X) to [out=-155,in=-110, looseness=2] (X);
			\draw[<->,>=latex, dashed] (Y) to [out=-25,in=-70, looseness=2] (Y);
			\draw[<->,>=latex, dashed] (W) to [out=-155,in=-110, looseness=2] (W);
			\draw[<->,>=latex, dashed] (Z) to [out=-25,in=-70, looseness=2] (Z);			
		\end{tikzpicture}
		\caption{}
		\label{fig:identifiable_macro:c}
	\end{subfigure}
    	\hfill 
	\begin{subfigure}{.16\textwidth}
		\centering
		\begin{tikzpicture}[{black, circle, draw, inner sep=0}]
			\tikzset{nodes={draw,rounded corners},minimum height=0.6cm,minimum width=0.6cm}	
			\tikzset{latent/.append style={white, fill=black}}
			
			\node[fill=red!30] (X) at (0,0) {$C_{\mathbb{X}}$} ;
            \node[fill=red!30] (Xprime) at (0,-1.2) {$C_{\mathbb{Z}}$} ;

            \node (R) at (1.4,1.2) {$C_{\mathbb{R}}$};
            \node (Z) at (2.4,0) {$C_{\mathbb{U}}$};
			\node (W) at (1,0.2) {$C_{\mathbb{W}}$};
			\node[fill=blue!30] (Y) at (1.8,-1.2) {$C_{\mathbb{Y}}$};

				\draw [->,>=latex] (Xprime) -- (X);

			\draw [->,>=latex,] (Xprime) -- (Y);
			\draw [<-,>=latex,] (X) -- (W);

			\draw[<-,>=latex] (W) -- (Xprime);
			\draw[->,>=latex] (W) -- (Z);
            
			 \draw[->,>=latex] (Y) -- (Z);
             \draw[->,>=latex] (W) -- (Y);
             \draw[->,>=latex] (W) -- (R);
             \draw[->,>=latex] (Z) -- (R);
             \draw[->,>=latex] (R) -- (Y);
             \draw[->,>=latex] (X) -- (R);

                \draw[<->,>=latex, dashed] (W) to [out=-105,in=145, looseness=1] (Y);

			
			\draw[->,>=latex] (X) to [out=165,in=120, looseness=2] (X);
                \draw[->,>=latex] (Xprime) to [out=165,in=120, looseness=2] (Xprime);
			\draw[->,>=latex] (Y) to [out=-25,in=-70, looseness=2] (Y);
			\draw[->,>=latex] (W) to [out=15,in=60, looseness=2] (W);
			\draw[->,>=latex] (Z) to [out=15,in=60, looseness=2] (Z);
			
			\draw[<->,>=latex, dashed] (Xprime) to [out=-155,in=-110, looseness=2] (Xprime);
			\draw[<->,>=latex, dashed] (X) to [out=-155,in=-110, looseness=2] (X);
			\draw[<->,>=latex, dashed] (Y) to [out=-155,in=-110, looseness=2] (Y);
			\draw[<->,>=latex, dashed] (W) to [out=165,in=120, looseness=2] (W);
			\draw[<->,>=latex, dashed] (Z) to [out=-25,in=-70, looseness=2] (Z);			
		\end{tikzpicture}
		\caption{}
		\label{fig:identifiable_macro:d}
	\end{subfigure}
	\caption{C-DMGs with identifiable macro causal effects. Each pair of red and blue vertices represents the causal effect we are interested in.}
	\label{fig:identifiable_macro}
\end{figure}

\citep{Pearl_1995} introduced an important tool in causal reasoning referred to as the do-calculus.
This do-calculus consists of three rules each relying on some d-separation in the ADMG to guarantee an equality between different probabilities.
Every one of these three rules can be interpreted differently: the first one allows the insertion or deletion of an observation, the second one allows for the exchange between actions and observations and the third one allows the insertion or deletion of actions.
The do-calculus in ADMGs is complete and thus allows, whenever it is possible, to identify causal effects.
In other words, it allows, whenever it is possible, to express a causal effect containing a $\interv{\cdot}$ operator as a probabilistic expression without any $\interv{\cdot}$ operator and thus allows to compute it from a positive observational distribution.
Since the do-calculus was initially introduced for ADMGs, it is not easily extendable to cyclic graphs. 
However, \cite{Forre_2020} showed that replacing d-separation by $\sigma$-separation in the three rules allows the do-calculus to be applicable on DMGs induced by ioSCMs.
In this subsection, we show that this version of the do-calculus  is also readily applicable to C-DMGs over DMGs.

There exists multiple equivalent ways of writing the do-calculus rules, for example \citep{Pearl_2000} uses the notion of mutilated graph to define the three rules of do-calculus.
In this paper, we will follow the notation used in~\cite{Forre_2020}.
To do so, we will extend every graph $\mathcal{G}^*=(\mathbb{V}^*, \mathbb{E}^*)$, by adding an intervention vertex $I_X$ and the edge $I_X \rightarrow X$ for every vertex of the graph $X\in\mathbb{V}^*$.
Moreover, we will use the $\sigma$-separation notation with a $\interv{\cdot}$ operator (\eg, $\sigsepcgraph{A}{B}{C,\interv{D}}{\mathcal{G}^*}$) to place ourselves in the intervened graph \ie, where all arrows going in $D$ are deleted and in which $D$ is conditioned on\footnote{In other words we write $\sigsepcgraph{A}{B}{C,\interv{D}}{\mathcal{G}^*}$ to mean, in \cite{Pearl_2000}'s notation, $\sigsepcgraph{A}{B}{C,D}{\mathcal{G}^*_{\overline{D}}}$ where $\mathcal{G}^*_{\overline{D}}$ is obtained from $\mathcal{G}^*$ by removing every edge going in $D$.}.
Using these newly defined notations and $\sigma$-separation we can now state the rules of do-calculus and show their applicability to C-DMGs over DMGs.

\begin{theorem}[do-calculus for C-DMGs over DMGs and macro causal effects]
	\label{theorem:soundness_do-calculus}
	Let $\mathcal{G}^\mathbb{c}=(\mathbb{C},\mathbb{E}^\mathbb{c})$ be a C-DMG over DMGs and $\mathbb{C}_\mathbb{X},\mathbb{C}_\mathbb{Y},\mathbb{C}_\mathbb{Z},\mathbb{C}_\mathbb{W}$ be disjoint subsets of $\mathbb{C}$.
The three following rules of the do-calculus are sound.
	\begin{equation*}
		\begin{aligned}
			\textbf{Rule 1:}& \probac{\mathbb{c}_{\mathbb{y}}}{\interv{\mathbb{c}_{\mathbb{z}}},\mathbb{c}_{\mathbb{x}},\mathbb{c}_{\mathbb{w}}} = \probac{\mathbb{c}_{\mathbb{y}}}{\interv{\mathbb{c}_{\mathbb{z}}},\mathbb{c}_{\mathbb{w}}}
			&\text{if } \sigsepcgraph{\mathbb{C}_{\mathbb{Y}}}{\mathbb{C}_{\mathbb{X}}}{\mathbb{C}_{\mathbb{W}},\interv{\mathbb{C}_{\mathbb{Z}}}}{\mathcal{G}^{\mathbb{c}}}\\
			\textbf{Rule 2:}& \probac{\mathbb{c}_{\mathbb{y}}}{\interv{\mathbb{c}_{\mathbb{z}}},\interv{\mathbb{c}_{\mathbb{x}}},\mathbb{c}^{\mathbb{w}}} = \probac{\mathbb{c}_{\mathbb{y}}}{\interv{\mathbb{c}_{\mathbb{z}}},\mathbb{c}_{\mathbb{x}},\mathbb{c}_{\mathbb{w}}}
			&\text{if } \sigsepcgraph{\mathbb{C}_{\mathbb{Y}}}{\mathbb{I}_{\mathbb{C}_{\mathbb{X}}}}{\mathbb{C}_{\mathbb{X}}, \mathbb{C}_{\mathbb{W}}, \interv{\mathbb{C}_{\mathbb{Z}}}}{\mathcal{G}^{\mathbb{c}}}\\
			\textbf{Rule 3:}& \probac{\mathbb{c}_{\mathbb{y}}}{\interv{\mathbb{c}_{\mathbb{z}}},\interv{\mathbb{c}_{\mathbb{x}}},\mathbb{c}_{\mathbb{w}}} = \probac{\mathbb{c}_{\mathbb{y}}}{\interv{\mathbb{c}_{\mathbb{z}}},\mathbb{c}_{\mathbb{w}}}
			&\text{if } \sigsepcgraph{\mathbb{C}_{\mathbb{Y}}}{\mathbb{I}_{\mathbb{C}_{\mathbb{X}}}}{\mathbb{C}_{\mathbb{W}},\interv{\mathbb{C}_{\mathbb{Z}}}}{\mathcal{G}^{\mathbb{c}}}\\
		\end{aligned}
	\end{equation*}
\end{theorem}

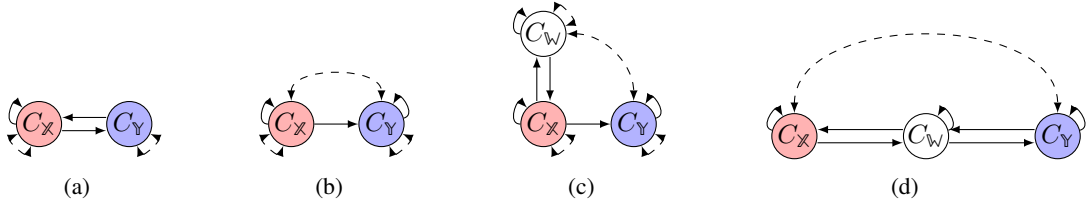
\begin{figure}[t!]
	\centering
	\begin{subfigure}{.15\textwidth}
		\centering
		\begin{tikzpicture}[{black, circle, draw, inner sep=0}]
			\tikzset{nodes={draw,rounded corners},minimum height=0.6cm,minimum width=0.6cm}	
			
			\node[fill=red!30] (X) at (0,0) {$C_{\mathbb{X}}$} ;
			\node[fill=blue!30] (Y) at (1.2,0) {$C_{\mathbb{Y}}$};
			
			\begin{scope}[transform canvas={yshift=-.25em}]
				\draw [->,>=latex] (X) -- (Y);
			\end{scope}
			\begin{scope}[transform canvas={yshift=.25em}]
				\draw [<-,>=latex] (X) -- (Y);
			\end{scope}
			
			\draw[->,>=latex] (X) to [out=180,in=135, looseness=2] (X);

            \draw[<->,>=latex, dashed] (X) to [out=-155,in=-110, looseness=2] (X);
        \draw[<->,>=latex, dashed] (Y) to [out=-25,in=-70, looseness=2] (Y);

		\end{tikzpicture}
		\caption{\centering}
		\label{fig:non_identifiable_macro:a}
	\end{subfigure}
	\hfill 
	\begin{subfigure}{.15\textwidth}
		\centering
		\begin{tikzpicture}[{black, circle, draw, inner sep=0}]
			\tikzset{nodes={draw,rounded corners},minimum height=0.6cm,minimum width=0.6cm}	
			
			\node[fill=red!30] (X) at (0,0) {$C_{\mathbb{X}}$} ;
			\node[fill=blue!30] (Y) at (1.2,0) {$C_{\mathbb{Y}}$};
			
			\draw [->,>=latex] (X) -- (Y);
			\draw[<->,>=latex, dashed] (X) to [out=90,in=90, looseness=1] (Y);
			
			\draw[->,>=latex] (X) to [out=180,in=135, looseness=2] (X);
			\draw[->,>=latex] (Y) to [out=15,in=60, looseness=2] (Y);

    			\draw[<->,>=latex, dashed] (X) to [out=-155,in=-110, looseness=2] (X);
			\draw[<->,>=latex, dashed] (Y) to [out=-25,in=-70, looseness=2] (Y);

		\end{tikzpicture}
		\caption{\centering}
		\label{fig:non_identifiable_macro:b}
	\end{subfigure}
	\hfill 
	\begin{subfigure}{.15\textwidth}
		\centering
		\begin{tikzpicture}[{black, circle, draw, inner sep=0}]
			\tikzset{nodes={draw,rounded corners},minimum height=0.6cm,minimum width=0.6cm}	
			\tikzset{anomalous/.append style={fill=easyorange}}
			\tikzset{rc/.append style={fill=easyorange}}
			
			\node (W) at (0,1.2) {$C_{\mathbb{W}}$} ;
			\node[fill=red!30] (X) at (0,0) {$C_{\mathbb{X}}$} ;
			\node[fill=blue!30] (Y) at (1.2,0) {$C_{\mathbb{Y}}$};
			
			\draw [->,>=latex] (X) -- (Y);
			\begin{scope}[transform canvas={xshift=-.25em}]
				\draw [->,>=latex] (X) -- (W);
			\end{scope}
			\begin{scope}[transform canvas={xshift=.25em}]
				\draw [<-,>=latex] (X) -- (W);
			\end{scope}
			\draw[<->,>=latex, dashed] (W) to [out=0,in=90, looseness=1] (Y);
			
			\draw[->,>=latex] (W) to [out=180,in=135, looseness=2] (W);
			\draw[->,>=latex] (X) to [out=180,in=135, looseness=2] (X);
			\draw[->,>=latex] (Y) to [out=15,in=60, looseness=2] (Y);

  			\draw[<->,>=latex, dashed] (X) to [out=-25,in=-70, looseness=2] (X);
			\draw[<->,>=latex, dashed] (Y) to [out=-25,in=-70, looseness=2] (Y);
			\draw[<->,>=latex, dashed] (W) to [out=20,in=65, looseness=2] (W);			

		\end{tikzpicture}
		\caption{\centering}
		\label{fig:non_identifiable_macro:c}
	\end{subfigure}
	\hfill 
	\begin{subfigure}{.28\textwidth}
            \centering
            \begin{tikzpicture}[{black, circle, draw, inner sep=0}]
                    \tikzset{nodes={draw,rounded corners},minimum height=0.6cm,minimum width=0.6cm}	
                    \tikzset{latent/.append style={white, fill=black}}
        
                    \node[fill=red!30] (X) at (0,0) {$C_{\mathbb{X}}$} ;
                    \node[fill=blue!30] (Y) at (3.5,0) {$C_{\mathbb{Y}}$};
                    \node (Z) at (1.75,0) {$C_{\mathbb{W}}$};

         \begin{scope}[transform canvas={yshift=-.25em}]
             \draw [->,>=latex] (X) -- (Z);
             \end{scope}
         \begin{scope}[transform canvas={yshift=.25em}]
             \draw [<-,>=latex] (X) -- (Z);
             \end{scope}			
        
         \begin{scope}[transform canvas={yshift=-.25em}]
             \draw [->,>=latex] (Z) -- (Y);
             \end{scope}
         \begin{scope}[transform canvas={yshift=.25em}]
             \draw [<-,>=latex] (Z) -- (Y);
             \end{scope}

                    \draw[<->,>=latex, dashed] (X) to [out=90,in=90, looseness=1] (Y);
        
                    \draw[->,>=latex] (X) to [out=165,in=120, looseness=2] (X);
                    \draw[->,>=latex] (Y) to [out=15,in=60, looseness=2] (Y);
                    \draw[->,>=latex] (Z) to [out=15,in=60, looseness=2] (Z);
        
                \end{tikzpicture}
            \caption{}
            \label{fig:non_identifiable_macro:d}
        \end{subfigure}
	\caption{C-DMGs with not identifiable macro causal effects. Each pair of red and blue vertices represents the total effect we are interested in.}
	\label{fig:non_identifiable_macro}
\end{figure}%

Now that the rules of do-calculus has been stated for C-DMGs over DMGs and their soundness proven, one can use them sequentially to identify the causal effect $\probac{\mathbb{c}_\mathbb{y}}{\interv{\mathbb{c}_\mathbb{x}}}$ in all C-DMGs over DMGs in Figure~\ref{fig:identifiable_macro}.
That is to say, to express the causal effect as a probabilistic expression without any $\interv{\cdot}$ operator.
This is done in the following examples.

\begin{example}
Both in Figure~\ref{fig:identifiable_macro:a} and~\ref{fig:identifiable_macro:c}, one can verify that $\sigsepcgraph{C_{\mathbb{Y}}}{I_{C_{\mathbb{X}}}}{C_{\mathbb{X}}}{\mathcal{G}^{\mathbb{c}}}$, thus Rule 2 of the do-calculus is applicable and $\probac{c_{\mathbb{y}}}{\interv{c_{\mathbb{x}}}} = \probac{c_{\mathbb{y}}}{c_{\mathbb{x}}}$.
\end{example}

\begin{example}
 Notice that Figure~\ref{fig:identifiable_macro:b} does not contain any cycle other than self-loops and is very similar to Figure 1(b) of \cite{Anand_2023} which corresponds to the well-known front-door criterion~\citep{Pearl_2000}. \citep{Forre_2018} have shown that in the acyclic case, $\sigma$-separation coincides with d-separation. Thus, using the corresponding sequence of classical rules of probability and rules of do-calculus as the one given in \citep[p.83]{Pearl_2000}, one obtains $\probac{c_{\mathbb{y}}}{\interv{c_{\mathbb{x}}}} = \sum_{c_{\mathbb{w}}} \probac{c_{\mathbb{w}}}{c_{\mathbb{x}}} \sum_{c_{\mathbb{x}'}} \probac{c_{\mathbb{y}}}{c_{\mathbb{w}},c_{\mathbb{x}'}} \proba{c_{\mathbb{x}'}}$.\end{example}

\begin{example}
Consider the C-DMG over DMGs in Figure~\ref{fig:identifiable_macro:d} containing a cycle between $C_{\mathbb{Y}}$, $C_{\mathbb{R}}$, and $C_{\mathbb{U}}$ and a hidden confounding between  
$C_{\mathbb{Y}}$ and $C_{\mathbb{W}}$. Let 
$\probac{c_{\mathbb{y}}}{\interv{c_{\mathbb{x}},c_{\mathbb{z}}}}$ be the causal effect of interest. Using the rule of total probability we can rewrite $\probac{c_{\mathbb{y}}}{\interv{c_{\mathbb{x}},c_{\mathbb{z}}}}$ as
$$\sum_{c_{\mathbb{w}}} \probac{c_{\mathbb{y}}}{\interv{c_{\mathbb{x}},c_{\mathbb{z}}},c_{\mathbb{w}}} \probac{c_{\mathbb{w}}}{\interv{c_{\mathbb{x}},c_{\mathbb{z}}}}.$$

We first focus on $\probac{c_{\mathbb{y}}}{\interv{c_{\mathbb{x}},c_{\mathbb{z}}},c_{\mathbb{w}}}$.
Notice that $\sigsepcgraph{C_{\mathbb{Y}}}{I_{C_{\mathbb{X}}}}{C_{\mathbb{X}}, C_{\mathbb{W}},\interv{C_{\mathbb{Z}}}}{\mathcal{G}^{\mathbb{c}}}$  
and that $\sigsepcgraph{C_{\mathbb{Y}}}{I_{C_{\mathbb{Z}}}}{C_{\mathbb{Z}}, C_{\mathbb{X}}, C_{\mathbb{W}}}{\mathcal{G}^{\mathbb{c}}}$ 
which means using two consecutive Rule 2  we can rewrite $\probac{c_{\mathbb{y}}}{\interv{c_{\mathbb{x}},c_{\mathbb{z}}},c_{\mathbb{w}}}$ as:
$\probac{c_{\mathbb{y}}}{c_{\mathbb{z}},c_{\mathbb{x}}, c_{\mathbb{w}}}.$

Now we focus on $\probac{c_{\mathbb{w}}}{\interv{c_{\mathbb{x}},c_{\mathbb{z}}}}$.
Notice that $\sigsepcgraph{C_{\mathbb{W}}}{\mathbb{I}_{C_{\mathbb{X}}}}{\interv{C_{\mathbb{Z}}}}{\mathcal{G}^{\mathbb{c}}}$  which means by Rule 3 of the do-calculus we can completely remove $\interv{c_{\mathbb{x}}}$ from the expression.
Furthermore, we have $\sigsepcgraph{C_{\mathbb{W}}}{\mathbb{I}_{C_{\mathbb{Z}}}}{C_{\mathbb{Z}}}{\mathcal{G}^{\mathbb{c}}}$ which means by Rule 2 we can replace $\interv{c_{\mathbb{z}}}$ by $c_{\mathbb{z}}$.
So we can rewrite $\probac{c_{\mathbb{w}}}{\interv{c_{\mathbb{x}},c_{\mathbb{z}}}}$ as $\probac{c_{\mathbb{w}}}{c_{\mathbb{z}}}$.
\end{example}

These three examples show how the rules of do-calculus in C-DMGs on DMGs can be used to write macro causal effects as an expression of observed probabilities.
Therefore, the macro causal effects in these C-DMGs can be estimated from the observational data, provided there is no further issues in the data (\eg, positivity violations).

In the following theorem, we show that not only is the do-calculus applicable in C-DMGs over DMGs, but it is also complete.

\begin{theorem}[Completeness of do-calculus for C-DMGs and macro causal effects]
	\label{theorem:completeness_do-calculus}
If one of the do-calculus rules does not apply for a given C-DMG over DMGs, then there exists a compatible DMG for which the corresponding rule does not apply.
\end{theorem}

Note that this completeness result links C-DMGs to the underlying compatible DMGs, however it does not guarantee the absence of other rules to identify causal effects.
While the do-calculus based on d-separation in ADMGs introduced in \citep{Pearl_1995} has been proven to be complete~\citep{Shpitser_2006, Huang_2006}, the do-calculus based on $\sigma$-separation in DMGs is not yet proven to be complete and thus our results suffer the same limitation.
Using the completeness of the do-calculus in C-DMGs (Theorem~\ref{theorem:completeness_do-calculus}), one can be convinced, by going through every possible sequence of rules of the do-calculus, that the causal effects of interest in every C-DMGs depicted in Figure~\ref{fig:DAG_ADMG_CDMG} and~\ref{fig:non_identifiable_macro} is not identifiable.
Unfortunately, going through every possible sequence of rules of the do-calculus is time-consuming and can become impractical when considering larger graphs.
In an effort to solve this issue, the following subsection introduces a sub-graphical structure which allows to recognize more efficiently when a causal effect is not identifiable using the do-calculus.


\subsection{Non-Identifiability: a graphical characterization}

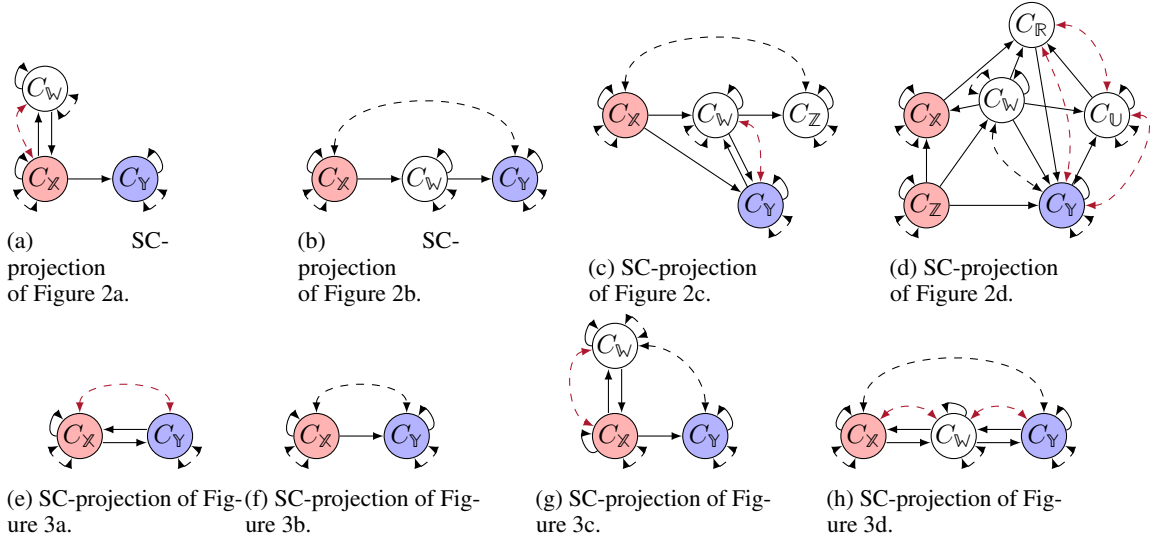
\begin{figure}[t!]
	\centering
    	\begin{subfigure}{.15\textwidth}
		\centering
		\begin{tikzpicture}[{black, circle, draw, inner sep=0}]
			\tikzset{nodes={draw,rounded corners},minimum height=0.6cm,minimum width=0.6cm}	
			\tikzset{anomalous/.append style={fill=easyorange}}
			\tikzset{rc/.append style={fill=easyorange}}
			
			\node (W) at (0,1.2) {$C_{\mathbb{W}}$} ;
			\node[fill=red!30] (X) at (0,0) {$C_{\mathbb{X}}$} ;
			\node[fill=blue!30] (Y) at (1.2,0) {$C_{\mathbb{Y}}$};
			
			\draw [->,>=latex] (X) -- (Y);
			\begin{scope}[transform canvas={xshift=-.25em}]
				\draw [->,>=latex] (X) -- (W);
			\end{scope}
			\begin{scope}[transform canvas={xshift=.25em}]
				\draw [<-,>=latex] (X) -- (W);
			\end{scope}
			
			\draw[->,>=latex] (W) to [out=180,in=135, looseness=2] (W);
			\draw[->,>=latex] (X) to [out=180,in=135, looseness=2] (X);
			\draw[->,>=latex] (Y) to [out=15,in=60, looseness=2] (Y);

			\draw[<->,>=latex, dashed, sorbonnered] (W) to [out=220,in=120, looseness=1] (X);

			\draw[<->,>=latex, dashed] (X) to [out=-155,in=-110, looseness=2] (X);
			\draw[<->,>=latex, dashed] (Y) to [out=-25,in=-70, looseness=2] (Y);
			\draw[<->,>=latex, dashed] (W) to [out=-15,in=-60, looseness=2] (W);	
		\end{tikzpicture}
		\caption{SC-projection of  Figure~\ref{fig:identifiable_macro:a}.}
		\label{fig:proj_identifiable:a}
	\end{subfigure}
	\hfill  
	\begin{subfigure}{.15\textwidth}
		\centering
		\begin{tikzpicture}[{black, circle, draw, inner sep=0}]
			\tikzset{nodes={draw,rounded corners},minimum height=0.6cm,minimum width=0.6cm}	
			\tikzset{latent/.append style={white, fill=black}}
			
			\node[fill=red!30] (X) at (0,0) {$C_{\mathbb{X}}$} ;
			\node[fill=blue!30] (Y) at (2.4,0) {$C_{\mathbb{Y}}$};
			\node (Z) at (1.2,0) {$C_{\mathbb{W}}$};

			\draw [->,>=latex,] (X) -- (Z);

			\draw[->,>=latex] (Z) -- (Y);

			\draw[<->,>=latex, dashed] (X) to [out=90,in=90, looseness=1] (Y);
			
			\draw[->,>=latex] (X) to [out=165,in=120, looseness=2] (X);
			\draw[->,>=latex] (Y) to [out=15,in=60, looseness=2] (Y);
			\draw[->,>=latex] (Z) to [out=15,in=60, looseness=2] (Z);

			\draw[<->,>=latex, dashed] (X) to [out=-155,in=-110, looseness=2] (X);
			\draw[<->,>=latex, dashed] (Y) to [out=-25,in=-70, looseness=2] (Y);
			\draw[<->,>=latex, dashed] (Z) to [out=-25,in=-70, looseness=2] (Z);			
			
		\end{tikzpicture}
		\caption{SC-projection of  Figure~\ref{fig:identifiable_macro:b}.}
		\label{fig:proj_identifiable:b}
	\end{subfigure}
	\hfill 
	\begin{subfigure}{.16\textwidth}
		\centering
		\begin{tikzpicture}[{black, circle, draw, inner sep=0}]
			\tikzset{nodes={draw,rounded corners},minimum height=0.6cm,minimum width=0.6cm}	
			\tikzset{latent/.append style={white, fill=black}}
			
			\node[fill=red!30] (X) at (0,0) {$C_{\mathbb{X}}$} ;
			\node (Z) at (2.4,0) {$C_{\mathbb{Z}}$};
			\node (W) at (1.2,0) {$C_{\mathbb{W}}$};
			\node[fill=blue!30] (Y) at (1.8,-1.2) {$C_{\mathbb{Y}}$};

			\draw [->,>=latex,] (X) -- (Y);
			\draw [->,>=latex,] (X) -- (W);

			\draw[->,>=latex] (W) -- (Z);
         \begin{scope}[transform canvas={yshift=-.1em, xshift=-.1em}]
				\draw [->,>=latex] (Y) -- (W);
			\end{scope}
			\begin{scope}[transform canvas={yshift=.1em, xshift=.1em}]
				\draw [<-,>=latex] (Y) -- (W);
			\end{scope}			
			
			\draw[<->,>=latex, dashed] (X) to [out=90,in=90, looseness=1] (Z);
			\draw[<->,>=latex, dashed, sorbonnered] (W) to [out=-15,in=90, looseness=1] (Y);
			
			\draw[->,>=latex] (X) to [out=165,in=120, looseness=2] (X);
			\draw[->,>=latex] (Y) to [out=15,in=60, looseness=2] (Y);
			\draw[->,>=latex] (W) to [out=15,in=60, looseness=2] (W);
			\draw[->,>=latex] (Z) to [out=15,in=60, looseness=2] (Z);
			
			\draw[<->,>=latex, dashed] (X) to [out=-155,in=-110, looseness=2] (X);
			\draw[<->,>=latex, dashed] (Y) to [out=-25,in=-70, looseness=2] (Y);
			\draw[<->,>=latex, dashed] (W) to [out=-155,in=-110, looseness=2] (W);
			\draw[<->,>=latex, dashed] (Z) to [out=-25,in=-70, looseness=2] (Z);			
		\end{tikzpicture}
		\caption{SC-projection of  Figure~\ref{fig:identifiable_macro:c}.}
		\label{fig:proj_identifiable:c}
	\end{subfigure}
	\hfill 
	\begin{subfigure}{.16\textwidth}
		\centering
		\begin{tikzpicture}[{black, circle, draw, inner sep=0}]
			\tikzset{nodes={draw,rounded corners},minimum height=0.6cm,minimum width=0.6cm}	
			\tikzset{latent/.append style={white, fill=black}}
			
			\node[fill=red!30] (X) at (0,0) {$C_{\mathbb{X}}$} ;
            \node[fill=red!30] (Xprime) at (0,-1.2) {$C_{\mathbb{Z}}$} ;

            \node (R) at (1.4,1.2) {$C_{\mathbb{R}}$};
            \node (Z) at (2.4,0) {$C_{\mathbb{U}}$};
			\node (W) at (1,0.2) {$C_{\mathbb{W}}$};
			\node[fill=blue!30] (Y) at (1.8,-1.2) {$C_{\mathbb{Y}}$};

				\draw [->,>=latex] (Xprime) -- (X);

			\draw [->,>=latex,] (Xprime) -- (Y);
			\draw [<-,>=latex,] (X) -- (W);

			\draw[<-,>=latex] (W) -- (Xprime);
			\draw[->,>=latex] (W) -- (Z);
            
			 \draw[->,>=latex] (Y) -- (Z);
             \draw[->,>=latex] (W) -- (Y);
             \draw[->,>=latex] (W) -- (R);
             \draw[->,>=latex] (Z) -- (R);
             \draw[->,>=latex] (R) -- (Y);
             \draw[->,>=latex] (X) -- (R);

                \draw[<->,>=latex, dashed] (W) to [out=-105,in=145, looseness=1] (Y);

            \draw[<->,>=latex, dashed, sorbonnered] (R) to [out=0,in=90, looseness=1] (Z);
            \draw[<->,>=latex, dashed, sorbonnered] (Z) to [out=0,in=0, looseness=1] (Y);
            \draw[<->,>=latex, dashed, sorbonnered] (R) to [out=-65,in=80, looseness=1] (Y);
			
			\draw[->,>=latex] (X) to [out=165,in=120, looseness=2] (X);
                \draw[->,>=latex] (Xprime) to [out=165,in=120, looseness=2] (Xprime);
			\draw[->,>=latex] (Y) to [out=15,in=60, looseness=2] (Y);
			\draw[->,>=latex] (W) to [out=15,in=60, looseness=2] (W);
			\draw[->,>=latex] (Z) to [out=15,in=60, looseness=2] (Z);
			
			\draw[<->,>=latex, dashed] (Xprime) to [out=-155,in=-110, looseness=2] (Xprime);
			\draw[<->,>=latex, dashed] (X) to [out=-155,in=-110, looseness=2] (X);
			\draw[<->,>=latex, dashed] (Y) to [out=-25,in=-70, looseness=2] (Y);
			\draw[<->,>=latex, dashed] (W) to [out=165,in=120, looseness=2] (W);
			\draw[<->,>=latex, dashed] (Z) to [out=-25,in=-70, looseness=2] (Z);			
		\end{tikzpicture}
		\caption{SC-projection of  Figure~\ref{fig:identifiable_macro:d}.}
		\label{fig:proj_identifiable:d}
	\end{subfigure}

	\begin{subfigure}{.22\textwidth}
		\centering
		\begin{tikzpicture}[{black, circle, draw, inner sep=0}]
			\tikzset{nodes={draw,rounded corners},minimum height=0.6cm,minimum width=0.6cm}	
			\tikzset{anomalous/.append style={fill=easyorange}}
			\tikzset{rc/.append style={fill=easyorange}}
			
			\node[fill=red!30] (X) at (0,0) {$C_{\mathbb{X}}$} ;
			\node[fill=blue!30] (Y) at (1.2,0) {$C_{\mathbb{Y}}$};
			
			\begin{scope}[transform canvas={yshift=-.25em}]
				\draw [->,>=latex] (X) -- (Y);
			\end{scope}
			\begin{scope}[transform canvas={yshift=.25em}]
				\draw [<-,>=latex] (X) -- (Y);
			\end{scope}
			
			\draw[->,>=latex] (X) to [out=180,in=135, looseness=2] (X);

			\draw[<->,>=latex, dashed] (X) to [out=-155,in=-110, looseness=2] (X);
			\draw[<->,>=latex, dashed] (Y) to [out=-25,in=-70, looseness=2] (Y);

   		\draw[<->,>=latex, dashed, sorbonnered] (X) to [out=90,in=90, looseness=1] (Y);

		\end{tikzpicture}
		\caption{SC-projection of  Figure~\ref{fig:non_identifiable_macro:a}.}
		\label{fig:proj_non_identifiable:a}
	\end{subfigure}
    \begin{subfigure}{.22\textwidth}
		\centering
		\begin{tikzpicture}[{black, circle, draw, inner sep=0}]
			\tikzset{nodes={draw,rounded corners},minimum height=0.6cm,minimum width=0.6cm}	
			\tikzset{anomalous/.append style={fill=easyorange}}
			\tikzset{rc/.append style={fill=easyorange}}
			
			\node[fill=red!30] (X) at (0,0) {$C_{\mathbb{X}}$} ;
			\node[fill=blue!30] (Y) at (1.2,0) {$C_{\mathbb{Y}}$};
			
			\draw [->,>=latex] (X) -- (Y);
			\draw[<->,>=latex, dashed] (X) to [out=90,in=90, looseness=1] (Y);
			
			\draw[->,>=latex] (X) to [out=180,in=135, looseness=2] (X);
			\draw[->,>=latex] (Y) to [out=15,in=60, looseness=2] (Y);

  			\draw[<->,>=latex, dashed] (X) to [out=-155,in=-110, looseness=2] (X);
			\draw[<->,>=latex, dashed] (Y) to [out=-25,in=-70, looseness=2] (Y);

		\end{tikzpicture}
		\caption{SC-projection of  Figure~\ref{fig:non_identifiable_macro:b}.}
		\label{fig:proj_non_identifiable:b}
	\end{subfigure}
	\hfill 
	\begin{subfigure}{.22\textwidth}
		\centering
		\begin{tikzpicture}[{black, circle, draw, inner sep=0}]
			\tikzset{nodes={draw,rounded corners},minimum height=0.6cm,minimum width=0.6cm}	
			\tikzset{anomalous/.append style={fill=easyorange}}
			\tikzset{rc/.append style={fill=easyorange}}
			
			\node (W) at (0,1.2) {$C_{\mathbb{W}}$} ;
			\node[fill=red!30] (X) at (0,0) {$C_{\mathbb{X}}$} ;
			\node[fill=blue!30] (Y) at (1.2,0) {$C_{\mathbb{Y}}$};
			
			\draw [->,>=latex] (X) -- (Y);
			\begin{scope}[transform canvas={xshift=-.25em}]
				\draw [->,>=latex] (X) -- (W);
			\end{scope}
			\begin{scope}[transform canvas={xshift=.25em}]
				\draw [<-,>=latex] (X) -- (W);
			\end{scope}
			\draw[<->,>=latex, dashed] (W) to [out=0,in=90, looseness=1] (Y);
			
			\draw[->,>=latex] (W) to [out=180,in=135, looseness=2] (W);
			\draw[->,>=latex] (X) to [out=220,in=175, looseness=2] (X);
			\draw[->,>=latex] (Y) to [out=15,in=60, looseness=2] (Y);

			\draw[<->,>=latex, dashed] (X) to [out=-25,in=-70, looseness=2] (X);
			\draw[<->,>=latex, dashed] (Y) to [out=-25,in=-70, looseness=2] (Y);
			\draw[<->,>=latex, dashed] (W) to [out=20,in=65, looseness=2] (W);			

   			\draw[<->,>=latex, dashed, sorbonnered] (W) to [out=190,in=155, looseness=1] (X);

		\end{tikzpicture}
		\caption{SC-projection of  Figure~\ref{fig:non_identifiable_macro:c}.}
		\label{fig:proj_non_identifiable:c}
	\end{subfigure}
        \hfill
         	\begin{subfigure}{.22\textwidth}
		\centering
		\begin{tikzpicture}[{black, circle, draw, inner sep=0}]
			\tikzset{nodes={draw,rounded corners},minimum height=0.6cm,minimum width=0.6cm}	
			\tikzset{latent/.append style={white, fill=black}}
			
			\node[fill=red!30] (X) at (0,0) {$C_{\mathbb{X}}$} ;
			\node[fill=blue!30] (Y) at (2.4,0) {$C_{\mathbb{Y}}$};
			\node (Z) at (1.2,0) {$C_{\mathbb{W}}$};

			\begin{scope}[transform canvas={yshift=-.25em}]
				\draw [->,>=latex] (X) -- (Z);
			\end{scope}
			\begin{scope}[transform canvas={yshift=.25em}]
				\draw [<-,>=latex] (X) -- (Z);
			\end{scope}			
			
			\begin{scope}[transform canvas={yshift=-.25em}]
				\draw [->,>=latex] (Z) -- (Y);
			\end{scope}
			\begin{scope}[transform canvas={yshift=.25em}]
				\draw [<-,>=latex] (Z) -- (Y);
			\end{scope}

			\draw[<->,>=latex, dashed] (X) to [out=90,in=90, looseness=1] (Y);
			
			\draw[->,>=latex] (X) to [out=165,in=120, looseness=2] (X);
			\draw[->,>=latex] (Y) to [out=15,in=60, looseness=2] (Y);
			\draw[->,>=latex] (Z) to [out=60,in=105, looseness=2] (Z);
			
			\draw[<->,>=latex, dashed] (X) to [out=-155,in=-110, looseness=2] (X);
			\draw[<->,>=latex, dashed] (Y) to [out=-25,in=-70, looseness=2] (Y);
			\draw[<->,>=latex, dashed] (Z) to [out=-25,in=-70, looseness=2] (Z);			
   			\draw[<->,>=latex, dashed, sorbonnered] (X) to [out=45,in=135, looseness=1] (Z);
   			\draw[<->,>=latex, dashed, sorbonnered] (Z) to [out=45,in=135, looseness=1] (Y);

		\end{tikzpicture}
		\caption{SC-projection of  Figure~\ref{fig:non_identifiable_macro:d}.}
		\label{fig:proj_non_identifiable:d}
	\end{subfigure}
	\caption{SC-projections of the C-DMGs in Figures~\ref{fig:DAG_ADMG_CDMG}, \ref{fig:identifiable_macro}, and Figures~\ref{fig:non_identifiable_macro}. Each pair of red and blue vertices represents the total effect we are interested in, and the red edges indicate those added through the SC-projection.}
	\label{fig:projections}
\end{figure}%

In ADMGs, there exists a sub-graphical structure, called a hedge~\citep{Shpitser_2006}, which is employed to  graphically characterize non-identifiability as shown in~\citealt[Theorem 4]{Shpitser_2006}.
This graphical criterion is complete when considering ADMGs, that is to say, if the total effect of $X$ on $Y$ is not identifiable in an ADMG then there exists a hedge for $(X,Y)$~\citep{Shpitser_2006}.  
However, it has been shown that this characterization is too weak to characterize all non-identifiabilities in the case of C-DMGs over ADMGs~\citep{Ferreira_2025b}.
Thus, a modified version of the hedge structure called the SC-hedge (strongly connected hedge) has been introduced in~\cite{Ferreira_2025}.
The presence of such SC-hedge in C-DMGs over ADMGs guarantees non-identifiability~\citep{Ferreira_2025,Ferreira_2025b}.
In this subsection, we will show that SC-hedge can also be used to characterize non identifiability in C-DMGs over DMGs in specific conditions.
First, let us recall some useful concepts to define a hedge.

\begin{definition}[C-component,  \cite{Tian_2002}]
    Let $\mathcal{G}^*=(\mathbb{V}^*, \mathbb{E}^*)$ be a graph.
    A subset of vertices $\mathbb{V}^*_C \subseteq \mathbb{V}^*$ such that $\forall V^*_1,V^*_n \in \mathbb{V}^*_C,~\exists V^*_1,\cdots,V^*_n\in \mathbb{V}^*$ with $\forall 1\leq i < n,~V^*_i\longdashleftrightarrow V^*_{i+1}$ is called a C-component.
\end{definition}
\begin{definition}[C-forest, \cite{Shpitser_2006}]
    Let $\mathcal{G}^*=(\mathbb{V}^*, \mathbb{E}^*)$ be a graph.
    If $\mathcal{G}^*$ is acyclic, $\mathcal{G}^*$ is a forest (\ie, every of its vertices has at most one child), and $\mathcal{G}^*$ is a C-component then $\mathcal{G}^*$ is called a C-forest.
    The vertices which have no children are called roots and we say a C-forest is $\mathbb{R}^*$-rooted if it has roots $\mathbb{R}^*\subseteq\mathbb{V}^*$.
\end{definition}

\begin{definition}[Hedge,  \cite{Shpitser_2006,Shpitser_2008}]
Consider a graph $\mathcal{G}^*=(\mathbb{V}^*, \mathbb{E}^*)$ and two disjoint sets of vertices $\mathbb{X}^*, \mathbb{Y}^*\subseteq \mathbb{V}^*$.
Let $\mathcal{F}=(\mathbb{V}^*_\mathcal{F}, \mathbb{E}^*_\mathcal{F})$ and $\mathcal{F}'=(\mathbb{V}^*_{\mathcal{F}'}, \mathbb{E}^*_{\mathcal{F}'})$ be two $\mathbb{R}^*$-rooted C-forests subgraphs of $\mathcal{G}^*$ such that $\mathbb{X}^* \cap \mathbb{V}^*_{\mathcal{F}} \ne\emptyset$, $\mathbb{X}^* \cap \mathbb{V}^*_{\mathcal{F}'} =\emptyset$, $\mathbb{F}' \subseteq \mathbb{F}$, and $\mathbb{R}^*\subset \ancestors{\mathbb{Y}^*}{\mathcal{G}^*\backslash\mathbb{X}^*}$. Then $\mathbb{F}$ and $\mathbb{F}'$ form a hedge for the pair $(\mathbb{X}^*, \mathbb{Y}^*)$ in $\mathcal{G}^*$.
\end{definition}

As mentioned before, a hedge turned out to be too weak to cover non-identifiability in C-DMGs \citep{Ferreira_2025b}.
For example, the C-DMG in Figure~\ref{fig:non_identifiable_macro:a} contains no hedge but the macro causal effect is not identifiable due to the cycle between $\mathbb{C}_{\mathbb{X}}$ and $\mathbb{C}_{\mathbb{Y}}$.
In the following, we formally define SC-hedges in the context of C-DMGs over DMGs and demonstrate that this substructure serves as a sound criterion for detecting non-identifiable macro causal effects if every cluster in a cycle is of size strictly greater than 1.

\begin{definition}[Strongly connected projection (SC-projection)]
\label{def:SC-proj}
    Consider a C-DMG $\mathcal{G}^{\mathbb{c}}=(\mathbb{C}, \mathbb{E}^{\mathbb{c}})$.
    The SC-projection $\mathcal{H}^{\mathbb{c}}$ of $\mathcal{G}^{\mathbb{c}}$ is the graph that includes all vertices and edges from $\mathcal{G}^{\mathbb{c}}$, plus a dashed bidirected edge between each pair $C_\mathbb{X},C_\mathbb{Y}\in \mathbb{C}$ such that $\scc{C_\mathbb{X}}{\mathcal{G}^{\mathbb{c}}} = \scc{C_\mathbb{Y}}{\mathcal{G}^{\mathbb{c}}}$ and $C_\mathbb{X}\ne C_\mathbb{Y}$.
\end{definition}

\begin{definition}[Strongly Connected Hedge (SC-Hedge)]
\label{def:SC-Hedge}
    Consider an C-DMG $\mathcal{G}^{\mathbb{c}}=(\mathbb{C}, \mathbb{E}^{\mathbb{c}})$, its SC-projection $\mathcal{H}^{\mathbb{c}}$ and two disjoints sets of vertices $\mathbb{C}_\mathbb{X},\mathbb{C}_\mathbb{Y}\subseteq\mathbb{C}$.
    A hedge for $(\mathbb{C}_\mathbb{X},\mathbb{C}_\mathbb{Y})$ in $\mathcal{H}^{\mathbb{c}}$ is an SC-hedge for $(\mathbb{C}_\mathbb{X},\mathbb{C}_\mathbb{Y})$ in $\mathcal{G}^\mathbb{c}$.
\end{definition}

The following theorem guarantees the soundness of the SC-hedge criterion in C-DMGs over DMGs when every cluster in a cycle is of size strictly greater than 1.
This assumption is useful as it allows the existence of compatible ADMGs even when the C-DMG over DMGs contains cycles.
Thus, one can use \citealt[Theorem 5]{Ferreira_2025b} in the presence of an SC-hedge to show that for every identifying sequence of do-calculus rules, there exists a compatible ADMG in which this sequence is not applicable.

\begin{theorem}
\label{theorem:SC-Hedge}
    Consider an C-DMG $\mathcal{G}^{\mathbb{c}}=(\mathbb{C}, \mathbb{E}^{\mathbb{c}})$ such that every cluster which is in a cycle is of size at least 2 and two disjoints sets of vertices $\mathbb{C}_\mathbb{X},\mathbb{C}_\mathbb{Y}\subseteq\mathbb{C}$.
    If there exists an SC-hedge for $(\mathbb{C}_\mathbb{X},\mathbb{C}_\mathbb{Y})$ in $\mathcal{G}^{\mathbb{c}}$ then $\probac{\mathbb{c}_{\mathbb{y}}}{\interv{\mathbb{c}_{\mathbb{x}}}}$ is not identifiable.
\end{theorem}

The SC-projections of the C-DMGs illustrated in Figures~\ref{fig:identifiable_macro} and \ref{fig:non_identifiable_macro} are given in Figure~\ref{fig:projections}. One can notice that the SC-projections of the C-DMGs in Figure~\ref{fig:non_identifiable_macro} all contain a hedge, thus the C-DMGs in Figure~\ref{fig:non_identifiable_macro} contain a SC-hedge and the causal effect of interest is therefore not identifiable according to Theorem~\ref{theorem:SC-Hedge}. In contrast, the projections of the C-DMGs in Figure~\ref{fig:identifiable_macro} do not contain a hedge, this highlights the usefulness of SC-hedges for causal effect identification.

\section{Conclusion}
\label{sec:conclusion}

In this paper, we established the soundness and completeness of $\sigma$-separation and the do-calculus using $\sigma$-separation for identifying macro causal effects in C-DMGs over DMGs. 
There are two main limitations to this work.
The first limitation is that the completeness result in Theorem~\ref{theorem:completeness_do-calculus} does not take into account that there might exist different sequences of rules of the do-calculus in different DMGs that can give the same final identification of the causal effect.
Moreover, the completeness results only link the cluster graphs to the compatible underlying graphs, our results say nothing on the completeness of $\sigma$-separation and do-calculus in DMGs.
A second related limitation is that we provided a graphical characterization for the non-identifiability of macro causal effects, however this characterization is not proven to be complete, even though we did not find any counter-example of its completeness.
Proving it to be complete remains an open problem.

\Myack{This work was supported by the CIPHOD project (ANR-23-CPJ1-0212-01).}

\bibliography{references}

\newpage 
\appendix

\section{Appendix}
\subsection{Additional Notations and Properties}

In order to map the vertices in a C-DMG $\mathcal{G}^\mathbb{c}$ with the vertices in a compatible DMG $\mathcal{G}$, we will use the notion of corresponding cluster: $\forall C \in \mathbb{C},~\forall V\in C,~\cluster{V}{\mathcal{G}^\mathbb{c}}=C$.
Additionally, we will write for every set of clusters $\mathbb{C}_\mathbb{A}\subseteq\mathbb{C},~\mathbb{A}=\bigcup_{C\in\mathbb{C}_\mathbb{A}}C$.

\begin{definition}[Maximal compatible DMG]
    \label{def:max_comp_DMG}
    Let $\mathcal{G}^\mathbb{c} = (\mathbb{C}, \mathbb{E}^\mathbb{c})$ be a C-DMG.
    Let us define the following sets:
	\begin{equation*}
		\begin{aligned}
			\mathbb{V} :=& \bigcup_{C\in\mathbb{C}}C&\\
            \mathbb{E}^\rightarrow :=& \{V \rightarrow V' \mid \forall V\in C,~V'\in C'\st C\rightarrow C'\in\mathbb{E}^\mathbb{c}\}&\\
            \mathbb{E}^{\longdashleftrightarrow} :=& \{V\longdashleftrightarrow V' \mid  \forall V\in C,~V'\in C'\st C\longdashleftrightarrow C'\in\mathbb{E}^\mathbb{c}\}&\\
            \mathbb{E} :=& \mathbb{E}^\rightarrow \cup \mathbb{E}^{\longdashleftrightarrow} &\\
		\end{aligned}
	\end{equation*}
    The graph $\maxcomp{\mathcal{G}^\mathbb{c}} = (\mathbb{V},\mathbb{E})$ is compatible with $\mathcal{G}^\mathbb{c}$ and for every compatible DMG $\mathcal{G}$, we have $\mathcal{G}\subseteq\maxcomp{\mathcal{G}^\mathbb{c}}$, thus $\maxcomp{\mathcal{G}^\mathbb{c}}$ is called the maximal compatible DMG of $\mathcal{G}^\mathbb{c}$.
\end{definition}

\begin{property}[Compatibility of extended graphs]
    \label{prop:compatibility_extended}
    Let $\mathcal{G}^\mathbb{c}$ be a C-DMG and $\mathcal{G}$ be a compatible DMG.
    Let us consider $\extended{\mathcal{G}^{\mathbb{c}}}$ and $\extended{\mathcal{G}}$ the corresponding extended graphs.
    Take $\extended{\mathbb{V}}=\mathbb{V}\cup\{I_V\mid V \in \mathbb{V}\}$ the extended micro variables and $\extended{\mathbb{C}}=\mathbb{C}\cup\{\{I_V\mid V \in C\} \mid C \in \mathbb{C}\}$ the extended partition.
    $\extended{\mathcal{G}}$ is compatible with $\extended{\mathcal{G}^{\mathbb{c}}}$ according to partition $\extended{\mathbb{C}}$.
\end{property}

\begin{proof}
    Firstly, $\forall V,V'\in\mathbb{V},~V\rightarrow V' \in \extended{\mathcal{G}} (\text{resp. }\longdashleftrightarrow) \iff V\rightarrow V' \in \mathcal{G} (\text{resp. }\longdashleftrightarrow) \iff \cluster{V}{\mathcal{G}^\mathbb{c}}\rightarrow\cluster{V'}{\mathcal{G}^\mathbb{c}} \in \mathcal{G}^\mathbb{c} (\text{resp. }\longdashleftrightarrow) \iff \cluster{V}{\mathcal{G}^\mathbb{c}}\rightarrow\cluster{V'}{\mathcal{G}^\mathbb{c}} \in \extended{\mathcal{G}^{\mathbb{c}}} (\text{resp. }\longdashleftrightarrow)$.
    Secondly, $\forall V\in\mathbb{V},~I_V\rightarrow V \in \extended{\mathcal{G}} \text{ and } I_{\cluster{V}{\mathcal{G}^\mathbb{c}}}\rightarrow\cluster{V}{\mathcal{G}^\mathbb{c}} \in \extended{\mathcal{G}^{\mathbb{c}}}$.
    Lastly, $\forall C\in\mathbb{C},~I_C\rightarrow C \in \extended{\mathcal{G}^{\mathbb{c}}} \text{ and } \exists V\in C,~I_V \rightarrow V \in \extended{\mathcal{G}}$ because $\mathbb{C}$ is a partition so $\forall C \in \mathbb{C},~C\neq \emptyset$.
\end{proof}

\begin{property}[Extended maximal compatible graphs]
    \label{prop:ext_max_comp}
    Let $\mathcal{G}^\mathbb{c}$ be a C-DMG, $\maxcomp{\mathcal{G}^\mathbb{c}}$ be the maximal compatible graph of $\mathcal{G}^\mathbb{c}$, $\extended{\maxcomp{\mathcal{G}^\mathbb{c}}}$ be the extended graph of $\maxcomp{\mathcal{G}^\mathbb{c}}$ and $\maxcomp{\extended{\mathcal{G}^{\mathbb{c}}}}$ be the maximal compatible graph of $\extended{\mathcal{G}^{\mathbb{c}}}$.
    These graphs verify
    \begin{itemize}
        \item $\extended{\maxcomp{\mathcal{G}^\mathbb{c}}}\subseteq\maxcomp{\extended{\mathcal{G}^{\mathbb{c}}}}$, and 
        \item $\forall V \in \extended{\mathbb{V}},~\scc{V}{\extended{\maxcomp{\mathcal{G}^\mathbb{c}}}}=\scc{V}{\maxcomp{\extended{\mathcal{G}^{\mathbb{c}}}}}$.
    \end{itemize}
    Therefore, any macro-level $\sigma$-connection that holds in $\maxcomp{\extended{\mathcal{G}^{\mathbb{c}}}}$ also holds in $\extended{\maxcomp{\mathcal{G}^\mathbb{c}}}$.
\end{property}

\begin{proof} 
    Firstly, $\forall V,V'\in\mathbb{V},~V\rightarrow V' \in \extended{\maxcomp{\mathcal{G}^\mathbb{c}}} (\text{resp. }\longdashleftrightarrow) \iff V\rightarrow V' \in \maxcomp{\mathcal{G}^\mathbb{c}} (\text{resp. }\longdashleftrightarrow) \iff \cluster{V}{\mathcal{G}^\mathbb{c}}\rightarrow\cluster{V'}{\mathcal{G}^\mathbb{c}} \in \mathcal{G}^\mathbb{c} (\text{resp. }\longdashleftrightarrow) \iff \cluster{V}{\mathcal{G}^\mathbb{c}}\rightarrow\cluster{V'}{\mathcal{G}^\mathbb{c}} \in \extended{\mathcal{G}^\mathbb{c}} (\text{resp. }\longdashleftrightarrow) \iff V\rightarrow V' \in \maxcomp{\extended{\mathcal{G}^{\mathbb{c}}}}(\text{resp. }\longdashleftrightarrow)$.
    Secondly, $\forall V\in\mathbb{V},~I_V\rightarrow V \in \extended{\maxcomp{\mathcal{G}^\mathbb{c}}} \text{ and } I_V \in \cluster{I_V}{\extended{\mathcal{G}^{\mathbb{c}}}},~V\in\cluster{V}{\extended{\mathcal{G}^{\mathbb{c}}}},~\cluster{I_V}{\extended{\mathcal{G}^{\mathbb{c}}}}\rightarrow\cluster{V}{\extended{\mathcal{G}^{\mathbb{c}}}}\in\extended{\mathcal{G}^{\mathbb{c}}} \text{ thus } I_V\rightarrow V \in \maxcomp{\extended{\mathcal{G}^{\mathbb{c}}}}$.

    Regarding the strongly connected components, $\forall V \in \mathbb{V},~\scc{V}{\extended{\maxcomp{\mathcal{G}^\mathbb{c}}}} = \scc{V}{\maxcomp{\mathcal{G}^\mathbb{c}}} = \{V\}\cup\left(\bigcup_{C \in \sccEx{\cluster{V}{\mathcal{G}^\mathbb{c}}}{\mathcal{G}^\mathbb{c}}}C\right)\footnote{While the usual strongly connected component, $\scc{V^*}{\mathcal{G}^*}$, always contains the vertex $V^*$ itself, this tweaked version, $\sccEx{V^*}{\mathcal{G}^*}$, contains the vertex $V^*$ if and only if there exists a self-loop on it \ie, $V^*\rightarrow V^*\in\mathcal{G}^*$.} = \{V\}\cup\left(\bigcup_{C \in \sccEx{\cluster{V}{\extended{\mathcal{G}^\mathbb{c}}}}{\extended{\mathcal{G}^\mathbb{c}}}}C\right) = \scc{V}{\maxcomp{\extended{\mathcal{G}^{\mathbb{c}}}}}$.
    Moreover, $\forall V \in \mathbb{V}$, the set of edges including $I_V$ in $\extended{\maxcomp{\mathcal{G}^\mathbb{c}}}$ is $\{I_V\rightarrow V\}$ and the set of edges including $I_V$ in $\maxcomp{\extended{\mathcal{G}^{\mathbb{c}}}}$ is $\{I_V\rightarrow V'\mid V'\in\cluster{V}{\mathcal{G}^\mathbb{c}}\}$ thus $\scc{I_V}{\extended{\maxcomp{\mathcal{G}^\mathbb{c}}}}=\{I_V\}=\scc{I_V}{\maxcomp{\extended{\mathcal{G}^{\mathbb{c}}}}}$.

    In conclusion, using the Definition~\ref{def:s-separation}, any macro-level $\sigma$-connection that holds in $\maxcomp{\extended{\mathcal{G}^{\mathbb{c}}}}$ also holds in $\extended{\maxcomp{\mathcal{G}^\mathbb{c}}}$.
\end{proof}

\begin{property}[Compatibility of intervened graphs]
    \label{prop:compatibility_intervened}
    Let $\mathcal{G}^\mathbb{c}$ be a C-DMG and $\mathcal{G}$ be a compatible DMG.
    Let $\mathbb{C}_\mathbb{A}\subseteq \mathbb{C}$ be a set of clusters.
    Let us consider $\intervened{\mathcal{G}^\mathbb{c}}{\mathbb{C}_\mathbb{A}}$ and $\intervened{\mathcal{G}}{\mathbb{A}}$ the graphs obtained respectively by intervening on $\mathbb{C}_\mathbb{A}$ in $\mathcal{G}^\mathbb{c}$ and by intervening on $\mathbb{A}$ in $\mathcal{G}$.
    $\intervened{\mathcal{G}}{\mathbb{A}}$ is compatible with $\intervened{\mathcal{G}^\mathbb{c}}{\mathbb{C}_\mathbb{A}}$.

    Moreover, let $\maxcomp{\mathcal{G}^\mathbb{c}}$ be the maximal compatible graph of $\mathcal{G}^\mathbb{c}$, $\intervened{\maxcomp{\mathcal{G}^\mathbb{c}}}{\mathbb{A}}$ its intervened graph and $\maxcomp{\intervened{\mathcal{G}^{\mathbb{c}}}{\mathbb{C}_\mathbb{A}}}$ be the maximal compatible graph of $\intervened{\mathcal{G}^{\mathbb{c}}}{\mathbb{C}_\mathbb{A}}$.
    $\intervened{\maxcomp{\mathcal{G}^\mathbb{c}}}{\mathbb{A}}$ and $\maxcomp{\intervened{\mathcal{G}^{\mathbb{c}}}{\mathbb{C}_\mathbb{A}}}$ are the same graph.
\end{property}

\begin{proof}
    $\forall V,V' \in \mathbb{V},~V\rightarrow V' \in \intervened{\mathcal{G}}{\mathbb{A}} (\text{resp. }\longdashleftrightarrow) \iff V\rightarrow V'\in\mathcal{G}(\text{resp. }\longdashleftrightarrow)\text{ and }V'\notin\mathbb{A} \iff \cluster{V}{\mathcal{G}^\mathbb{c}}\rightarrow\cluster{V'}{\mathcal{G}^\mathbb{c}} \in \mathcal{G}^\mathbb{c}(\text{resp. }\longdashleftrightarrow)\text{ and }\cluster{V'}{\mathcal{G}^\mathbb{c}} \notin \mathbb{C}_\mathbb{A} \iff \cluster{V}{\mathcal{G}^\mathbb{c}}\rightarrow\cluster{V'}{\mathcal{G}^\mathbb{c}} \in \intervened{\mathcal{G}^\mathbb{c}}{\mathbb{C}_\mathbb{A}}(\text{resp. }\longdashleftrightarrow)$.
\end{proof}

\begin{property}[Intervened maximal compatible graphs]
    \label{prop:int_max_comp}
    Let $\mathcal{G}^\mathbb{c}$ be a C-DMG, $\maxcomp{\mathcal{G}^\mathbb{c}}$ be the maximal compatible graph of $\mathcal{G}^\mathbb{c}$, $\intervened{\maxcomp{\mathcal{G}^\mathbb{c}}}{\mathbb{A}}$ be the intervened graph of $\maxcomp{\mathcal{G}^\mathbb{c}}$ and $\maxcomp{\intervened{\mathcal{G}^{\mathbb{c}}}{\mathbb{C}_\mathbb{A}}}$ be the maximal compatible graph of $\intervened{\mathcal{G}^{\mathbb{c}}}{\mathbb{C}_\mathbb{A}}$.
    $\intervened{\maxcomp{\mathcal{G}^\mathbb{c}}}{\mathbb{A}}$ and $\maxcomp{\intervened{\mathcal{G}^{\mathbb{c}}}{\mathbb{C}_\mathbb{A}}}$ are the same graph.
\end{property}

\begin{proof}
    $\forall V,V' \in \mathbb{V},~V\rightarrow V' \in \intervened{\maxcomp{\mathcal{G}^\mathbb{c}}}{\mathbb{A}} (\text{resp. }\longdashleftrightarrow)\iff V\rightarrow V' \in \maxcomp{\mathcal{G}^\mathbb{c}} (\text{resp. }\longdashleftrightarrow)\text{ and } V'\notin \mathbb{A} \iff \cluster{V}{\mathcal{G}^\mathbb{c}}\rightarrow\cluster{V'}{\mathcal{G}^\mathbb{c}} \in \mathcal{G}^\mathbb{c} (\text{resp. }\longdashleftrightarrow)\text{ and }\cluster{V'}{\mathcal{G}^\mathbb{c}} \notin \mathbb{C}_\mathbb{A} \iff \cluster{V}{\mathcal{G}^\mathbb{c}}\rightarrow\cluster{V'}{\mathcal{G}^\mathbb{c}} \in\intervened{\mathcal{G}^{\mathbb{c}}}{\mathbb{C}_\mathbb{A}}(\text{resp. }\longdashleftrightarrow)\iff V\rightarrow V' \in \maxcomp{\intervened{\mathcal{G}^{\mathbb{c}}}{\mathbb{C}_\mathbb{A}}}(\text{resp. }\longdashleftrightarrow)$. 
\end{proof}

\subsection{Proof of Theorem~\ref{theorem:soundness_s-sep}}
\begin{proof}
	Suppose $\mathbb{C}_\mathbb{X}$ and $\mathbb{C}_\mathbb{Y}$ are $\sigma$-separated by $\mathbb{C}_\mathbb{W}$ in $\mathcal{G}^\mathbb{c}$ and there exists a compatible DMG $\mathcal{G}=(\mathbb{V},\mathbb{E})$ and a walk $\pi=\langle V_1,\cdots,V_n\rangle$ in $\mathcal{G}$ from $V_1 \in {\mathbb{X}}$ to $V_n \in {\mathbb{Y}}$ which is not $\sigma$-blocked by ${\mathbb{W}}$.
	Consider the walk $\tilde{\pi}=\langle C_1,\cdots,C_n\rangle$ with $\forall 1\leq i \leq n, C_i=\cluster{V_i}{\mathcal{G}^\mathbb{c}}$ and $\forall 1\leq i <n, \langle C_i\rightarrow C_{i+1}\rangle \subseteq \tilde{\pi}$ (resp. $\leftarrow, \longdashleftrightarrow$) $\iff \langle V_i\rightarrow V_{i+1}\rangle \subseteq \pi$ (resp. $\leftarrow, \longdashleftrightarrow$).
    $\tilde{\pi}$ is a walk from $\mathbb{C}_\mathbb{X}$ to $\mathbb{C}_\mathbb{Y}$ in $\mathcal{G}^\mathbb{c}$.
    Since $\mathbb{C}_\mathbb{X}$ and $\mathbb{C}_\mathbb{Y}$ are $\sigma$-separated by $\mathbb{C}_\mathbb{W}$, we know that $\mathbb{C}_\mathbb{W}$ $\sigma$-blocks $\tilde{\pi}$.    \begin{itemize}
        \item If $C_1\in\mathbb{W}$ or $C_n\in\mathbb{C}_\mathbb{W}$, then $V_1\in\mathbb{W}$ or $V_n\in\mathbb{W}$ and thus $\pi$ is $\sigma$-blocked by $\mathbb{W}$ which contradicts the initial assumption.
    \end{itemize}
    Otherwise, take $1<i<n$ such that $\langle C_{i-1},C_i,C_{i+1}\rangle$ is $\mathbb{C}_\mathbb{W}$-$\sigma$-blocked.
    \begin{itemize}
        \item If $\langle C_{i-1}\stararrow C_{i}\arrowstar C_{i+1}\rangle \subseteq \tilde{\pi}$ and $C_{i}\notin\mathbb{C}_\mathbb{W}$ then, $\langle V_{i-1}\stararrow V_{i}\arrowstar V_{i+1}\rangle \subseteq \pi$ and $V_{i}\notin \mathbb{W}$. Thus $\pi$ is $\sigma$-blocked by $\mathbb{W}$ which contradicts the initial assumption.
        
        \item If $\langle C_{i-1}\leftarrow C_{i}\arrowstar C_{i+1}\rangle \subseteq \tilde{\pi}$ and $C_{i}\in\mathbb{C}_\mathbb{W}\backslash\scc{C_{i-1}}{\mathcal{G}^\mathbb{c}}$ then, $\langle V_{i-1}\leftarrow V_{i}\arrowstar V_{i+1}\rangle \subseteq \pi$.
        Moreover, $\scc{V_{i-1}}{\mathcal{G}}\subseteq\bigcup_{C\in\scc{C_{i-1}}{\mathcal{G}^\mathbb{c}}}C$ and thus $V_i \in \mathbb{W}\backslash\scc{V_{i-1}}{\mathcal{G}}$.
        Therefore, $\pi$ is $\sigma$-blocked by $\mathbb{W}$ which contradicts the initial assumption.

        \item If $\langle C_{i-1}\stararrow C_{i}\rightarrow C_{i+1}\rangle \subseteq \tilde{\pi}$ and $C_{i}\in\mathbb{C}_\mathbb{W}\backslash\scc{C_{i+1}}{\mathcal{G}^\mathbb{c}}$ then, $\langle V_{i-1}\stararrow V_{i}\rightarrow V_{i+1}\rangle \subseteq \pi$.
        Moreover, $\scc{V_{i+1}}{\mathcal{G}}\subseteq\bigcup_{C\in\scc{C_{i+1}}{\mathcal{G}^\mathbb{c}}}C$ and thus $V_i \in \mathbb{W}\backslash\scc{V_{i+1}}{\mathcal{G}}$.
        Therefore, $\pi$ is $\sigma$-blocked by $\mathbb{W}$ which contradicts the initial assumption.

        \item If $\langle C_{i-1}\leftarrow C_{i}\rightarrow C_{i+1}\rangle \subseteq \tilde{\pi}$ and $C_{i}\in\mathbb{C}_\mathbb{W}\backslash\left(\scc{C_{i-1}}{\mathcal{G}^\mathbb{c}}\cap\scc{C_{i+1}}{\mathcal{G}^\mathbb{c}}\right)$ then, $\langle V_{i-1}\leftarrow V_{i}\rightarrow V_{i+1}\rangle \subseteq \pi$.
        Moreover, $\left(\scc{V_{i-1}}{\mathcal{G}}\cap\scc{V_{i+1}}{\mathcal{G}}\right)\subseteq\bigcup_{C\in\left(\scc{C_{i-1}}{\mathcal{G}^\mathbb{c}}\cap\scc{C_{i+1}}{\mathcal{G}^\mathbb{c}}\right)}C$ and thus $V_i \in \mathbb{W}\backslash\left(\scc{V_{i-1}}{\mathcal{G}}\cap\scc{V_{i+1}}{\mathcal{G}}\right)$.
        Therefore, $\pi$ is $\sigma$-blocked by $\mathbb{W}$ which contradicts the initial assumption.
    \end{itemize}
	In conclusion, the $\sigma$-separation is sound in C-DMGs over DMGs.
    
\end{proof}

\subsection{Proof of Theorem~\ref{theorem:completeness_s-sep}}

\begin{proof}
	Suppose $\mathbb{C}_\mathbb{X}$ and $\mathbb{C}_\mathbb{Y}$ are not $\sigma$-separated by $\mathbb{C}_\mathbb{W}$ in $\mathcal{G}^\mathbb{c}$.
	There exists an $\mathbb{C}_\mathbb{W}$-$\sigma$-active path $\pi=\langle C_1,\cdots,C_n\rangle$ with $C_1\in\mathbb{C}_\mathbb{X}$ and $C_n\in\mathbb{C}_\mathbb{Y}$.
    Take $\maxcomp{\mathcal{G}^\mathbb{c}}=(\mathbb{V},\mathbb{E})$ the maximal compatible DMG of $\mathcal{G}^\mathbb{c}$ as in Definition~\ref{def:max_comp_DMG}.
	Take for every cluster $C\in\mathbb{C}$ a representative of this cluster $V_C\in\mathbb{C}$.
	The maximal compatible graph $\maxcomp{\mathcal{G}^\mathbb{c}}$ contains the path $\maxcomp{\pi} = \langle V_{C_1}, \cdots, V_{C_n}\rangle$ and for every cluster $C\in\mathbb{C}$ and every variable in that cluster $V\in C$, $\scc{V}{\maxcomp{\mathcal{G}^\mathbb{c}}} = \bigcup_{C'\in\scc{C}{\mathcal{G}^\mathbb{c}}}C'$.
    Therefore, $\pi$ being $\mathbb{C}_\mathbb{W}$-$\sigma$-active in $\mathcal{G}^\mathbb{c}$ clearly implies that $\maxcomp{\pi}$ is $\mathbb{W}$-$\sigma$-active in $\maxcomp{\mathcal{G}^\mathbb{c}}$.
	In conclusion, the $\sigma$-separation criterion in C-DMGs over DMGs is complete.
    
    Notice, that not only did we prove Theorem~\ref{theorem:completeness_s-sep}\textemdash\ie,if a $\sigma$-separation does not hold in a C-DMG then there exists a compatible DMG in which the corresponding $\sigma$-separation does not hold\textemdash but we also explicitly exhibited this compatible DMG as being the maximal compatible DMG.
\end{proof}

\subsection{Proof of Theorem~\ref{theorem:soundness_do-calculus}}

\begin{proof}
	Let $\mathcal{G}^\mathbb{c}=(\mathbb{C},\mathbb{E}^\mathbb{c})$ be a C-DMG, $\mathcal{G}$ a compatible DMG and $\mathbb{C}_\mathbb{X},\mathbb{C}_\mathbb{Y},\mathbb{C}_\mathbb{Z},\mathbb{C}_\mathbb{W}\subseteq\mathbb{C}$ be disjoints subsets of vertices.
    Suppose a rule of the do-calculus applies in $\mathcal{G}^\mathbb{c}$ then Theorem~\ref{theorem:soundness_s-sep}, Property~\ref{prop:compatibility_extended} and Property~\ref{prop:compatibility_intervened} guarantees that this rule applies in $\mathcal{G}$.
    More explicitly:
    \begin{itemize}
        \item If rule $1$ applies \ie, $\sigsepcgraph{\mathbb{C}_{\mathbb{Y}}}{\mathbb{C}_{\mathbb{X}}}{\mathbb{C}_{\mathbb{W}},\interv{\mathbb{C}_{\mathbb{Z}}}}{\mathcal{G}^{\mathbb{c}}}$, then using Theorem~\ref{theorem:soundness_s-sep} as well as Properties~\ref{prop:compatibility_extended} and~\ref{prop:compatibility_intervened} one knows that $\sigsepcgraph{\mathbb{Y}}{\mathbb{X}}{\mathbb{W},\interv{\mathbb{Z}}}{\mathcal{G}}$ and thus rule $1$ applies in $\mathcal{G}$.
        \item If rule $2$ applies \ie, $\sigsepcgraph{\mathbb{C}_{\mathbb{Y}}}{\mathbb{I}_{\mathbb{C}_{\mathbb{X}}}}{\mathbb{C}_{\mathbb{X}}, \mathbb{C}_{\mathbb{W}}, \interv{\mathbb{C}_{\mathbb{Z}}}}{\mathcal{G}^{\mathbb{c}}}$, then using Theorem~\ref{theorem:soundness_s-sep} as well as Properties~\ref{prop:compatibility_extended} and~\ref{prop:compatibility_intervened} one knows that $\sigsepcgraph{\mathbb{Y}}{\mathbb{I}_{\mathbb{X}}}{\mathbb{X}, \mathbb{W}, \interv{\mathbb{Z}}}{\mathcal{G}}$ and thus rule $2$ applies in $\mathcal{G}$.
        \item If rule $3$ applies \ie, $\sigsepcgraph{\mathbb{C}_{\mathbb{Y}}}{\mathbb{I}_{\mathbb{C}_{\mathbb{X}}}}{\mathbb{C}_{\mathbb{W}},\interv{\mathbb{C}_{\mathbb{Z}}}}{\mathcal{G}^{\mathbb{c}}}$, then using Theorem~\ref{theorem:soundness_s-sep} as well as Properties~\ref{prop:compatibility_extended} and~\ref{prop:compatibility_intervened} one knows that $\sigsepcgraph{\mathbb{Y}}{\mathbb{I}_{\mathbb{X}}}{\mathbb{W},\interv{\mathbb{Z}}}{\mathcal{G}}$ and thus rule $3$ applies in $\mathcal{G}$.
    \end{itemize}    
    Notice that because $\mathbb{C}_\mathbb{X}$ and $\mathbb{C}_\mathbb{Z}$ are disjoint, the actions of taking the intervened graph and taking the extended graph can be done in any order without any repercussion in the $\sigma$-separations of interest.
    
	In conclusion, the do-calculus using $\sigma$-separation is sound in C-DMG over DMGs .
\end{proof}

\subsection{Proof of Theorem~\ref{theorem:completeness_do-calculus}}

\begin{proof}
	Let $\mathcal{G}^\mathbb{c}=(\mathbb{C},\mathbb{E}^\mathbb{c})$ be a C-DMG, $\maxcomp{\mathcal{G}^\mathbb{c}}$ be the maximal compatible DMG and $\mathbb{C}_\mathbb{X},\mathbb{C}_\mathbb{Y},\mathbb{C}_\mathbb{Z},\mathbb{C}_\mathbb{W}\subseteq\mathbb{C}$ be disjoints subsets of vertices.
	Suppose a rule of the do-calculus does not applies in $\mathcal{G}^\mathbb{c}$ then Theorem~\ref{theorem:completeness_s-sep}, Property~\ref{prop:compatibility_extended} and Property~\ref{prop:compatibility_intervened} show that this rule does not apply in $\maxcomp{\mathcal{G}^\mathbb{c}}$.
    More explicitly:
    \begin{itemize}
        \item If rule $1$ does not apply \ie, $\notsigsepcgraph{\mathbb{C}_{\mathbb{Y}}}{\mathbb{C}_{\mathbb{X}}}{\mathbb{C}_{\mathbb{W}},\interv{\mathbb{C}_{\mathbb{Z}}}}{\mathcal{G}^{\mathbb{c}}}$, then using Theorem~\ref{theorem:completeness_s-sep} as well as Properties~\ref{prop:ext_max_comp} and~\ref{prop:int_max_comp} one knows that $\notsigsepcgraph{\mathbb{Y}}{\mathbb{X}}{\mathbb{W},\interv{\mathbb{Z}}}{\maxcomp{\mathcal{G}^\mathbb{c}}}$ and thus rule $1$ does not apply in $\maxcomp{\mathcal{G}^\mathbb{c}}$.
        \item If rule $2$ does not apply \ie, $\notsigsepcgraph{\mathbb{C}_{\mathbb{Y}}}{\mathbb{I}_{\mathbb{C}_{\mathbb{X}}}}{\mathbb{C}_{\mathbb{X}}, \mathbb{C}_{\mathbb{W}}, \interv{\mathbb{C}_{\mathbb{Z}}}}{\mathcal{G}^{\mathbb{c}}}$, then using Theorem~\ref{theorem:completeness_s-sep} as well as Properties~\ref{prop:ext_max_comp} and~\ref{prop:int_max_comp} one knows that $\notsigsepcgraph{\mathbb{Y}}{\mathbb{I}_{\mathbb{X}}}{\mathbb{X}, \mathbb{W}, \interv{\mathbb{Z}}}{\maxcomp{\mathcal{G}^\mathbb{c}}}$ and thus rule $2$ does not apply in $\maxcomp{\mathcal{G}^\mathbb{c}}$.
        \item If rule $3$ does not apply \ie, $\notsigsepcgraph{\mathbb{C}_{\mathbb{Y}}}{\mathbb{I}_{\mathbb{C}_{\mathbb{X}}}}{\mathbb{C}_{\mathbb{W}},\interv{\mathbb{C}_{\mathbb{Z}}}}{\mathcal{G}^{\mathbb{c}}}$, then using Theorem~\ref{theorem:completeness_s-sep} as well as Properties~\ref{prop:ext_max_comp} and~\ref{prop:int_max_comp} one knows that $\notsigsepcgraph{\mathbb{Y}}{\mathbb{I}_{\mathbb{X}}}{\mathbb{W},\interv{\mathbb{Z}}}{\maxcomp{\mathcal{G}^\mathbb{c}}}$ and thus rule $3$ does not apply in $\maxcomp{\mathcal{G}^\mathbb{c}}$.
    \end{itemize}
    Notice that because $\mathbb{C}_\mathbb{X}$ and $\mathbb{C}_\mathbb{Z}$ are disjoint, considering the extended graph of the intervened graph or considering the intervened graph of the extended graph does not have any repercussion in the $\sigma$-connections of interest.
    
	In conclusion, the do-calculus using $\sigma$-separation is complete in C-DMG over DMGs.
\end{proof}

\subsection{Proof of Theorem~\ref{theorem:SC-Hedge}}

\begin{proof}
    Let $\mathcal{G}^{\mathbb{c}}=(\mathbb{C}, \mathbb{E}^{\mathbb{c}})$ be a C-DMG and take disjoint subsets $\mathbb{C}_\mathbb{X}, \mathbb{C}_\mathbb{Y} \subseteq \mathbb{C}$.
    Additionally, suppose that every cluster which is in a cycle in $\mathcal{G}^{\mathbb{c}}$ is of size at least 2.
    More formally, $\forall C\in\mathbb{C},~|\scc{C}{\mathcal{G}^{\mathbb{c}}}|>1\implies|C|>1$.
    Thanks to this assumption, one can view $\mathcal{G}^{\mathbb{c}}$ as a C-DMG over ADMGs and thus use prior work\citep{Ferreira_2025b}.
    Suppose there exists a SC-hedge for the pair $(\mathbb{C}_\mathbb{X}, \mathbb{C}_\mathbb{Y})$ in $\mathcal{G}^{\mathbb{c}}$.
    Then, according to Theorem 5 of \cite{Ferreira_2025b}, the effect of $\mathbb{C}_\mathbb{X}$ on $\mathbb{C}_\mathbb{Y}$ is not identifiable.

    In conclusion, the SC-hedge criterion is sound in C-DMG over DMGs under the additional assumption that every cluster which is in a cycle is of size at least 2.
\end{proof}

\end{document}